\documentclass[12pt]{article}
\usepackage[utf8]{inputenc}
\usepackage[english]{babel}
\usepackage{amssymb,amsmath,graphicx}
\usepackage{amsthm}
\usepackage{float}
\usepackage{afterpage}
\usepackage{dcolumn}
\usepackage[round,authoryear]{natbib}
\usepackage{varwidth}
\usepackage{enumitem}
 \usepackage{setspace}
\usepackage{cases}
\usepackage{bbm}
\usepackage{microtype}
\usepackage{multirow}
\usepackage{color}
\usepackage[ruled,linesnumbered]{algorithm2e}
\SetArgSty{textrm}
\SetAlFnt{\small}
\usepackage{natbib}
 \bibpunct[, ]{(}{)}{,}{a}{}{,}%
 \def\newblock{\ }%
\usepackage[hidelinks]{hyperref}
\newcommand{\SSe}{\mathcal S}

\allowdisplaybreaks

\vspace*{0.6cm}

\title{An Efficient Matheuristic for the Minimum-Weight  Dominating Set Problem}

\author{Mayra Albuquerque, Thibaut Vidal}

\begin{document}

\begin{center}

\begin{LARGE}
{An Efficient Matheuristic for the Minimum-Weight \vspace*{0.2cm}  \linebreak Dominating Set Problem}
\end{LARGE}

\vspace*{0.85cm}

\textbf{Mayra Albuquerque, Thibaut Vidal*} \\
Departamento de Informática, \\ 
Pontifícia Universidade Católica do Rio de Janeiro (PUC-Rio) \\
vidalt@inf.puc-rio.br \\

\vspace*{0.4cm}

August 2018 

Author Accepted Manuscript (AAM), published in \emph{Applied Soft Computing}

\url{https://doi.org/10.1016/j.asoc.2018.06.052}

\vspace*{0.15cm}

\end{center}
\noindent
\textbf{Abstract.}
A minimum dominating set in a graph is a minimum set of vertices such that every vertex of the graph either belongs to it, or is adjacent to one vertex of this set. This mathematical object is of high relevance in a number of applications related to social networks analysis, design of wireless networks, coding theory, and data mining, among many others. When vertex weights are given, minimizing the total weight of the dominating set gives rise to a problem variant known as the minimum weight dominating set problem. To solve this problem, we introduce a hybrid matheuristic combining a tabu search with an integer programming solver. The latter is used to solve subproblems in which only a fraction of the decision variables, selected relatively to the search history, are left free while the others are fixed.
Moreover, we introduce an adaptive penalty to promote the exploration of intermediate infeasible solutions during the search, enhance the algorithm with perturbations and node elimination procedures, and exploit richer neighborhood classes. Extensive experimental analyses on a variety of instance classes demonstrate the good performance of the algorithm, and the contribution of each component in the success of the search is analyzed.
\vspace*{0.3cm}


\vspace*{0.5cm}

\noindent
* Corresponding author

\newpage

\section{Introduction}

The minimum dominating set problem in a graph $G = (V,E)$ consists of determining a set $\mathcal{S} \subseteq V$ of minimum cardinality that dominates all vertices. A vertex is dominated if it belongs to $\mathcal{S}$ or has a neighbor in $\mathcal{S}$. In a variant of this problem, called Minimum Weight Dominating Set problem (MWDS), a non-negative weight is defined for each vertex, and the objective is to find a dominating set of minimum total weight. The MWDS is NP-hard \citep{Garey1990}, and the current exact methods cannot solve instances of practical relevance for data mining and other large-scale applications in reasonable CPU time.

Dominating set problems are linked with a rich set of applications \citep{Yu2013}, including the design of wireless sensor networks, the study of social networks and influence propagation \citep{Wang2011a,Wang2014,DaliriKhomami2018}, protein interaction networks \citep{Wuchty2014, Nacher2016} and covering codes \citep{Ostergard1997}, among others. Initially, a significant amount of research focused on approximation algorithms for this family of problems \citep[see, e.g., ][]{Chen2004, Zou2011, Schaudt2012}. In contrast, the research on efficient metaheuristics has expanded fairly recently. For the MWDS, some population-based algorithms have been introduced in \cite{Bouamama2016,Jovanovic2010} and \cite{Potluri2013}. The latter method has been successful for a wide range of problem instances thanks to its combination with an integer linear programming (ILP) solver. Finally, Wang et al. \cite{Wang2017a} introduced a local search with additional mechanisms to prevent cycling. These methods produce solutions of good quality on classical instance sets, however they may prematurely converge in some cases or tend to be over-restrictive, and require a significant amount of CPU time for the largest instances.

In this paper, we introduce a hybrid tabu search matheuristic (HTS-DS) for the MWDS. The term ``matheuristic'' has been widely used to refer to algorithms that combine metaheuristics with mixed integer programming (MIP) strategies and software \citep{Maniezzo2009,Ball2011,Archetti2014a}. The proposed method combines four successful strategies: an efficient neighborhood search, an adaptive penalty scheme to explore intermediate infeasible solutions, perturbation phases to promote exploration, and an intensification mechanism in the form of a MIP solver, which is applied to solve partial problems in which a fraction of the variables are fixed, keeping free those of the current best known solution and those associated to \emph{promising} vertices. As in the Construct, Merge, Solve {\&} Adapt (CMSA) approach described in \cite{Blum2016}, the set of promising vertices is selected based on the search history. Moreover, the size of the subproblem is adapted to exploit the capabilities of the MIP solver as efficiently as possible.

The performance of the proposed hybrid method is demonstrated through extensive experiments on a variety of benchmark instances with up to 4000 vertices and one million edges. The algorithm achieves solutions of better quality than previous methods, in a computational time which is notably smaller. Our sensitivity analyses on the method's components underline the decisive impact of some specific neighborhoods used in the tabu search, the perturbation mechanism, as well as the contribution of the integer programming solver. Finally, new best solutions have been produced for many classical benchmark instances.

The remainder of this paper is organized as follows. Section 2 formally defines the MWDS and reviews the related literature. Section 3 describes the proposed algorithm. Section 4 reports on our experimental analyses, and Section~5 concludes.

\section{Problem Statement and Literature Review}
\label{sec:statement}

A simple mathematical formulation of the MWDS is displayed in Equations (\ref{eq:fo}--\ref{eq:r2}).
In this formulation, the closed neighborhood $N(i)$ represents all vertices adjacent to $i$, including $i$ itself (i.e., $N(i) = \{i\} \cup \{j | (i,j)\in E\}$). Each decision variable $x_i$ is set to $1$ if vertex $i$ is included in the dominating set, and $0$ otherwise. The weight of each vertex $i$ is defined as~$w_i$.
\begin{align}
 \min \hspace{0.5cm}
 & \sum\limits_{i\in V} w_i x_i \label{eq:fo} \\
 \text{s.t.} \hspace{0.5cm}
 & \hspace{-0.2cm} \sum\limits_{k \in N(i)} x_{k} \geq 1 
 & \forall i \in V \label{eq:r1} \\
 & x_i \in \{0, 1\} & \forall i \in V \label{eq:r2} 
\end{align}

The MWDS can be viewed as a special case of the set covering (SC) problem, in which each vertex corresponds to a possible set. Although the research on exact methods has led to very efficient solution techniques for SC problems \citep{Caprara2000}, many instances of the MWDS present graphs with medium or high densities, leading to SC instances with large sets (i.e., dense matrices) which can be unusually challenging. For this reason, along with emerging applications in machine learning and social network analysis, a research line specific to the MWDS has been growing in intensity in recent years.

Earlier studies on the MWDS have focused on approximation algorithms for specific types of graphs called unit disk graphs (UDG), in relation to wireless ad-hoc networks applications. A UDG is a graph in which every vertex corresponds to a sensor on the plane, and in which two vertices are connected by an edge if their Euclidean distance in the plane is no more than $1$ unit.
The first constant-factor approximation algorithm for this setting was proposed by Ambuhl et al. \cite{Ambuhl2006}, with an approximation ratio of 72. Subsequently, this ratio has been successively improved, down to $(6+\epsilon)$ in  \cite{Huang2009} using a ``double-partition'' strategy,  followed by $(5+\epsilon)$ and $(4+\epsilon)$ in \cite{Dai2009} and \cite{Erlebach2010}. The first polynomial time approximation scheme (PTAS) was introduced in \cite{Zhu2012}, with an approximation ratio of $(1+\epsilon)$ for the specific case where the ratio of the weights of any two adjacent nodes is upper bounded by a constant. Finally, Li and Jin \cite{Li2015} introduced a PTAS for the general case without restrictions on adjacent~weights.

The research on metaheuristics for the MWDS has also recently grown. Many of the methods proposed for this problem rely on evolutionary population search. An ant colony algorithm with a pheromone correction strategy (Raka-ACO) was proposed in \cite{Jovanovic2010}. Subsequently, Potluri and Singh \cite{Potluri2013} introduced a hybrid genetic algorithm (HGA) and two extensions of the ACO algorithm with a local search which consists of removing redundant vertices. In the second algorithm, a pre-processing step is included immediately after pheromone initialization, in order to reinforce the pheromone values associated with 100 independent sets generated via a greedy algorithm. Later on, Lin et al. \cite{Lin2016} proposed a memetic algorithm based on a greedy randomised adaptive construction procedure as well as problem-tailored crossover and path-relinking operations.
An iterated greedy algorithm (R-PBIG) was introduced in \cite{Bouamama2016}. The method maintains a population of solutions and applies deconstruction and reconstruction steps. This approach was then hybridized with an ILP solver for solution improvement. Wang et al. \cite{Wang2017a} proposed a variant of tabu search called \emph{configuration checking} algorithm, with a mechanism which modifies the objective function based on the search history.
Finally, \cite{Chalupa2018} proposed a multi-start order-based randomised local search, using \emph{jump} moves for solution diversification, and reported computational results for a variety of MDS and MWDS instances.
Each of the mentioned methods led to some solution improvement on the classical benchmark instance sets for the problem. 
However, none of these methods reliably finds the best known or optimal solutions for all instances, and their computational time tends to be high for large graphs.
The contribution of this paper is to propose a scalable and efficient algorithm for the problem.

\section{Proposed Methodology}
\label{sec:methodology}

The proposed solution method, summarized in Algorithm \ref{pseudo-code}, combines a tabu search with an integer programming solver. The method starts from a random initial solution (Line~3). This solution is improved by a tabu search (Lines~3--18) with perturbation mechanisms (Line 15 -- after each $I_{\textsc{pert}}$ iterations), which terminates as soon as either $I_\textsc{NI}$ consecutive iterations (moves) without improvement of the best solution $\SSe_\textsc{best}$ or $I_\textsc{max}$ total iterations have been performed. The best solution of the tabu search then serves to define a reduced problem which is solved using an integer programming solver (Line~19). This process is repeated $N_\textsc{restart}$ times (Lines~2--22), and the best overall solution $\SSe_\textsc{overall}$ is returned (Line~23).

\begin{algorithm}[htb]
\linespread{1}\selectfont
$\SSe_\textsc{overall}  \gets \varnothing$ \tcp*{Stores the overall best solution}
 \For{$N_\textsc{restart}$ iterations}
{
$\SSe \gets$ Generate an initial feasible solution\;
 $\SSe_\textsc{best}  \gets \SSe$  \tcp*{Stores the best solution of current restart}
$i_\textsc{ni}  \gets 0$ \tcp*{Number of iterations without improvement}
 $i_\textsc{max} \gets 0$ \tcp*{Total number of iterations}
 \While{$i_\textsc{ni} < I_{\textsc{ni}}$ \textbf{and} $i_\textsc{max} < I_{\textsc{max}}$}
{
Update penalty factor $\alpha$ \;
$\SSe \gets$ Best non-tabu neighbor of $\SSe$ in \textsc{ADD},  \textsc{DEL} and  \textsc{SWAP}  \;
Update tabu list \;
\If{$\SSe$ is feasible \textbf{and} $f(\SSe) < f(\SSe_\textsc{best})$ }{$\SSe_\textsc{best} \gets \SSe$; $i_\textsc{ni} \gets 0$\; }
\If{$\exists \, k \in \mathbb Z_{> 0} \text{ such that }  i_\textsc{max} = k \times I_{\textsc{pert}}$} {$\SSe \gets \text{Perturbation} (\SSe_\textsc{best})$\;}
$i_\textsc{ni} \texttt{++}$;  
$ i_\textsc{max} \texttt{++}$\;
}
$\SSe \gets$ Construct and solve reduced IP\;
\lIf{$f(\SSe) < f(\SSe_\textsc{best})$ }{$\SSe_\textsc{best} \gets \SSe$}
\lIf{$f(\SSe_\textsc{best}) < f(\SSe_\textsc{overall})$ }{$\SSe_\textsc{overall} \gets \SSe_\textsc{best}$}
}
\textbf{return} $\SSe_\textsc{overall}$\;
 \caption{HTS-DS} \label{pseudo-code}
\end{algorithm}

We note that some penalized infeasible solutions, in which some vertices are not dominated, can be explored during the search. The objective function therefore plays an important role in the algorithm. This will be discussed in Section \ref{objective}. Subsequently, Section \ref{tabu} describes the solution initialization, tabu search and perturbation operator, and Section \ref{subproblem} presents the integer programming subproblem and its resolution.

\subsection{Penalized Objective Function}
\label{objective}

As illustrated in several previous studies (see, e.g., \cite{Glover2011,Vidal2013a}), an exploration of penalized infeasible solutions can help to diversify the search and prevent the method from becoming trapped in low-quality local optima. Therefore, HTS-DS relies on a penalized objective function that allows to evaluate infeasible solutions with some non-dominated vertices. The cost of a solution $\SSe$ is calculated as:
\begin{equation}
\label{fitnessW}
f(\SSe) = W(\SSe) + \alpha \times w^{\textsc{max}} \times N^\textsc{d}(\SSe),
\end{equation}
where $W(\SSe)$ is the total weight of solution $\SSe$, $\alpha$ is the current penalty factor, $w^{\textsc{max}}$ is the maximum weight of a vertex $i \in V$, and $N^\textsc{d}(\SSe)$ is the number of non-dominated vertices in solution $\SSe$. This way, the amount of penalty is directly proportional to the degree of infeasibility in order to promote a gradual return to the feasible solution~space.

Selecting a proper value for $\alpha$ is important for the efficiency of the search. However, in our preliminary experiments, a constant value was found to be insufficient for transitioning between different regions of the search space. We therefore designed a periodic ramp-up strategy, inspired by the strategic oscillation of \citep{Glover2011}, in which $\alpha$ rises from a minimum level $\alpha^{\textsc{min}}$ to a maximum level $\alpha^{\textsc{max}}$ by steps of $\alpha^{\textsc{step}}$, and then returns to its minimum value before increasing again. When $\alpha$ is small, the search will tend to remove vertices from the solution and lead to more non-dominated vertices. These vertices are subsequently covered again when the value of $\alpha$ becomes larger. Due to this behavior, HTS-DS also shares some common characteristics with ruin-and-recreate methods. With this analogy in mind, the parameter $\alpha^{\textsc{step}}$ has been set to be inversely proportional to the number of vertices of the graph,
\begin{equation}
\alpha^{\textsc{step}} = \frac{\alpha^{\textsc{max}} - \alpha^{\textsc{min}}}{\beta |V|}
\end{equation}
where $\beta$ is a parameter of the method which controls the number of steps between $\alpha^{\textsc{min}}$~and~$\alpha^{\textsc{max}}$, and the update rule (Line 8 of Algorithm 1) for parameter $\alpha$ is:
\begin{equation}
\alpha = 
\begin{cases}
\alpha + \alpha^{\textsc{step}} & \text{ if } \alpha < \alpha^{\textsc{max}} \\
\alpha^{\textsc{min}} &  \text{ otherwise.} 
\end{cases}
\end{equation}

\begin{figure}[htbp]
\includegraphics[width = \textwidth]{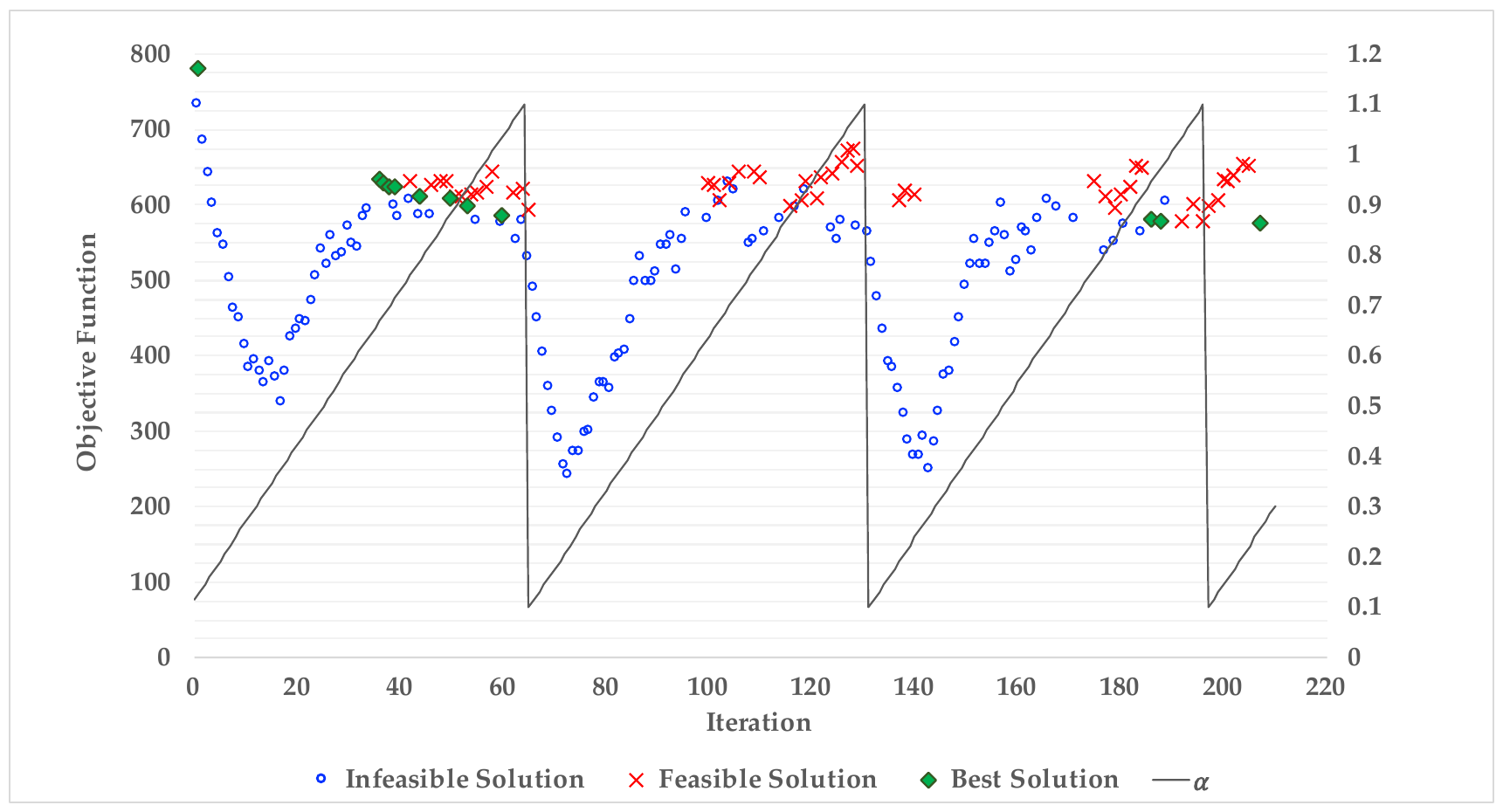} \vspace*{-1cm}
\caption{Evolution of the penalty parameter $\alpha$ during the search for a small instance (class SMPI -- problem $50\_50\_0$). Discovery of feasible, infeasible, and best solutions.}
\label{fig:objectivefunction}
\end{figure}

Figure \ref{fig:objectivefunction} illustrates the behavior of the parameter $\alpha$ on a small instance. We observe that the feasibility of the current solution directly depends on the value of~$\alpha$. The discovery of new best solutions, depicted with green diamonds in the graph, usually occurs when $\alpha$ is closer to $\alpha^{\textsc{max}}$, i.e., when the search is driven back towards feasible solutions. This controlled exploration of the infeasible solution space allows to transition towards structurally different solutions. 

\subsection{Tabu Search}
\label{tabu}

\paragraph{Initial Solution}
Each initial solution $\SSe$ of the tabu search is built by random construction (Line 3 of Algorithm 1). With uniform probability, the algorithm iteratively selects a random vertex which covers at least one non-dominated vertex, and inserts it in~$\SSe$. The process stops when a dominating set is obtained.

\paragraph{Neighborhood}
Solution $\SSe$ is then improved by tabu search considering three classical families of moves: 
\begin{itemize}[nosep]
\item  ADD -- adds a vertex $i$ into the solution;
\item  DEL -- removes a vertex $j$ from the solution;
\item  SWAP -- simultaneously adds a vertex $i$ and removes a vertex $j$ from the solution.
\end{itemize}
 
The moves are evaluated according to the objective function of Equation~(\ref{fitnessW}).
At each iteration, the best \emph{non-tabu} move from the entire neighborhood is applied (Line 9 of Algorithm 1). 
Note that, depending on the status of the tabu memory, this move can be deteriorating, thus allowing the algorithm to escape from local minima.

These three neighborhoods differ in their size: ADD and DEL contain~$\mathcal{O}(|V|)$ moves while SWAP contains $\mathcal{O}(|V|^2)$ moves. To balance the search effort and improve the speed of the method, we apply \emph{dynamic restrictions} to the SWAP neighborhood. This restriction works by first evaluating the ADD and DEL moves, ranking these moves relative to their impact on the solution value, and restricting the search of the SWAP neighborhoods to the $(i,j)$ pairs that belong to the top $\sqrt{|V|}$ ADD and DEL moves. With this restriction, only $\mathcal{O}(|V|)$ SWAP moves are evaluated. To efficiently evaluate each move, we maintain for each vertex $i$ a value $C(i)$ which counts how many times the vertex is dominated by another. Therefore, if $C(i) = 0$ then $i$ is non-dominated. This allows to evaluate each move with a complexity proportional to the degree of the associated vertex (or vertices).

\paragraph{Tabu List}
The short-term memory (tabu list) is essential to avoid cycling.
In HTS-DS, this list has a fixed size of $N_\textsc{Tabu}$ and contains two types of labels: those that prohibit the insertion of a specific vertex, and those that prohibit the removal of a specific vertex. The tabu list is updated (Line 10 of Algorithm 1) according to the following rules:
\begin{itemize}[nosep]
\item Whenever ADD($i$) is applied, a label is added to prohibit the removal of $i$;
\item Whenever DEL($j$) or  SWAP($i,j$) is applied, a label is added to prohibit the insertion~of~$j$.
\end{itemize}

Note that the tabu status associated to the SWAP move prohibits only the reinsertion of the removed vertex $j$. Indeed, in preliminary analyses, we observed that simultaneously prohibiting the removal of $i$ over-constrains the search and slows down the progress towards high-quality solutions. 
Finally, as usually done in most tabu searches, we use an aspiration criterion which revokes the tabu status of a move in case it leads to a new best solution.

\paragraph{Node Elimination}
Finally, after the application of each move, HTS-DS checks for the possible existence of \emph{redundant vertices}. A redundant vertex is a vertex which can be removed from the solution without increasing the number of non-dominated vertices. When such a situation occurs, a redundant vertex of maximum weight is removed. This process is iterated until no redundant vertex exists.

\paragraph{Perturbation}
After every $I_{\textsc{pert}}$ iterations of the tabu search, a perturbation mechanism is triggered to diversify the search further and reach different solutions (Line 15 of Algorithm 1). The perturbation is based on the ruin-and-recreate strategy \citep{Schrimpf2000}.
It works by removing  $\lfloor\rho\times|\SSe_\textsc{Best}|\rfloor$ vertices from the current best feasible solution $\SSe_\textsc{Best}$ of the tabu search, reconstructing the solution with a greedy algorithm, and finally applying the node elimination procedure. The resulting solution becomes the new starting point for the search.

The greedy algorithm used for solution reconstruction works as follows.
For every vertex~$i$, we define $\Gamma(i)$ as the set of non-dominated vertices belonging to $N(i)$ (adjacent to $i$, or $i$ itself). Let $\Delta(i) = |\Gamma(i)|$ and $W(i) = \sum_{k \in \Gamma_i} w_k$.
With equal probability, the greedy algorithm includes a vertex $i$ with:
\begin{itemize}[nosep]
\item highest value of $\Delta(i)/w_i$;
\item second-highest value of  $\Delta(i)/w_i$;
\item highest value of $W(i)/w_i$;
\item second-highest value of $W(i)/w_i$.
\end{itemize}

Ties (vertices with identical value) are broken randomly with uniform probability, and the process is iterated until a feasible solution is found.

\subsection{Resolution of a reduced integer problem}
\label{subproblem}

After the tabu search, HTS-DS constructs a reduced problem based on the information of the best current solution and the past search history, and solves it with an integer programming solver over the formulation of Equations (\ref{eq:fo}--\ref{eq:r2}) with the aim of finding a new best solution (Line 19 of Algorithm 1). In the reduced problem, most of the decision variables are fixed, and only a smaller group of \emph{free} variables remain, representing possible choices of vertices for the dominating set. This type of approach is classified as a \emph{decomposition} and \emph{partial optimization} method~in~\cite{Ball2011,Archetti2014a}.

The set of free variables corresponds to the $|\SSe_\textsc{Best}|$ vertices used in the best solution $\SSe_\textsc{best}$ of the tabu search, along with the $N_{\textsc{freq}}  = \max \{0,N_{\textsc{free}} - |\SSe_\textsc{Best}|\}$ \emph{most frequent} vertices observed in the incumbent solution, during the search history. Each other vertex is fixed by setting $x_i = 0$, i.e., excluding it from any candidate dominating set in the integer program.
To identify the most frequent vertices, HTS-DS counts the total number of iterations $I_\textsc{Total}(i)$ for which each vertex $i$ was used in the current solution, and selects the $N_{\textsc{freq}}$ vertices  with highest value of $I_\textsc{Total}(i)$. Such a counter can be efficiently implemented by storing the index of the current iteration when $i$ is included, and only updating the counter with the adequate increment when $i$ is removed.

The resulting formulation is solved by the integer programming solver subject to a time limit~$T_{\textsc{max}}$. Initially, the size parameter $N_{\textsc{free}}$ is set to $50$. Subsequently, in order to fully exploit the capabilities of the IP solver, the parameter $N_{\textsc{free}}$ is adapted from one general iteration of HTS-DS to the next. At the end of the resolution, there are three possible outcomes for the IP solver.
\begin{itemize}[nosep]
\item \emph{Case 1a) The solver finds an optimal solution, and $N_{\textsc{free}} = |V|$}. An optimal solution has been found for the MWDS problem, HTS-DS terminates.
\item \emph{Case 1b) The solver finds an optimal solution of the reduced problem.} In this case, the IP solver may be able to address a larger problem in the next iteration within the allowed time, and therefore the parameter $N_{\textsc{free}}$ is increased to $\min \{|V|,2 \times N_{\textsc{free}} \}$.
\item \emph{Case 2) The solution is not proven optimal or no solution is produced.} In this case, the IP solver has been used beyond its capabilities, and $N_{\textsc{free}}$ is reduced to $N_{\textsc{free}}/2$.
\end{itemize}
The best overall solution is stored and returned at the end of HTS-DS.

\section{Computational Experiments} 
\label{results}

Extensive computational experiments were conducted to evaluate the performance of the method relative to previous algorithms, and to examine the relative contribution of each of its main components. HTS-DS was implemented in C++, and CPLEX 12.7 was used for the resolution of the integer linear programs. All tests were conducted on a single thread of an I7-5820K 3.3GHz processor.

We rely on a total of 1060 problem instances originally proposed in \cite{Jovanovic2010}. These instances are divided into two types (T1 and T2) and two classes (SMPI and LPI). The SMPI class includes 320 small and medium instances with 50 to 250 vertices and 50 to 5000 edges, and the LPI class includes 210 larger instances counting between 300 and 1000 vertices, with up to 20000 edges. In the instances of type~T1, the vertex weights are uniformly distributed in the interval $[20, 70]$, while in the instances of type T2, the weight of each vertex $i$ is randomly chosen in $[1, {\delta(i)}^2]$, where $\delta(i)$ is the degree of $i$. Ten instances were generated for each problem dimension. Therefore, in the subsequent sections, all results will be aggregated (averaged) by groups of ten instances with the same number of edges and vertices.
All instance files and solutions are made available at the following address: \url{https://w1.cirrelt.ca/~vidalt/en/research-data.html}.

\subsection{Parameters Calibration}
\label{parameters}

Three main parameters of HTS-DS have a strong influence on the search: the size of the tabu list $N_\textsc{tabu}$, the strength of the perturbation operator $\rho$, and the control parameters for the penalty $(\alpha_\textsc{min},\alpha_\textsc{max},\beta)$.
 These parameters were calibrated through preliminary experiments by varying one parameter at a time until reaching a ``local optimum'' in terms of parameter configuration. Then, from the final parameter setting obtained, we performed a sensitivity analysis to examine the effect of a variation of each parameter. This analysis will be reported in Section \ref{sensitivity} along with other results measuring the contribution of the main search components.

Finally, the parameters $(N_\textsc{restart},I_{\textsc{max}},I_{\textsc{ni}},I_{\textsc{pert}},T_{\textsc{max}})$ control the number of iterations and search time of each component. Changing these parameters leads to different trade-offs between solution quality and computational effort. Therefore, for a fair experimental analysis, their values were selected to obtain solutions in a CPU time which is comparable to or lower than those of previous algorithms. The final parameter values are presented in Table \ref{tab:parameters}.

\begin{table}[htpb]
\centering
\renewcommand{\arraystretch}{1.1}
\setlength{\tabcolsep}{7pt}
\caption{Parameter configuration of HTS-DS}
\label{tab:parameters}
\scalebox{0.9}
{
\begin{tabular}{|l|l|r|}  
\hline  
\multicolumn{1}{|c|}{\textbf{Parameter}} & \multicolumn{1}{c|}{\textbf{Symbol}} & \multicolumn{1}{c|}{\textbf{Value}} \\ \hline
Number of restarts & $N_\textsc{restart}$ & 10 \\
Maximum number of iterations of TS & $I_{\textsc{max}}$ & 20000\\
Maximum number of iterations without improvement of TS & $I_{\textsc{ni}}$ & 10000\\
Frequency of the perturbation & $I_{\textsc{pert}}$ & 100\\
Perturbation strength & $\rho$ & 0.2\\
Size of the tabu list  & $N_\textsc{tabu}$ & 12\\
Time limit for the IP solver & $T_{\textsc{max}}$ & 1 sec\\ \hline
\multirow{3}{*}{Penalty factor} 
 & $\alpha^{\textsc{min}}$ & 0.1 \\
 & $\alpha^{\textsc{max}} $ & 1.1 \\
 & $\beta$    & 1.3\\ \hline
\end{tabular}  
}
\end{table}

 \subsection{Performance of the proposed algorithm}
 
HTS-DS was run ten times with different random seeds on each instance. We compare its results with those of the recent state-of-the-art algorithms for the MWDS listed in Table \ref{table:factor}. This table also indicates the CPU model used by each study, along with the associated time scaling factor (based on the \emph{Passmark} benchmark) representing the ratio between its speed and the speed of our processor.
In some cases, the CPU model was not reported by the authors, and therefore we used a factor of 1.0.
In the remainder of this section, the time values reported by previous studies will be multiplied by the associated factors, in order to account for CPU differences and conduct a fair comparison.

\begin{table}[htbp]
\renewcommand{\arraystretch}{1.2}
\caption{List of methods considered in the experiments, and CPU model information}
\label{table:factor}
\hspace*{-1.3cm}
\scalebox{0.9}
{
\begin{tabular}{|l|l|c|c|}
\hline
\multicolumn{1}{|c|}{\textbf{Acronym}} & \multicolumn{1}{c|}{\textbf{Description}} & \multicolumn{1}{c|}{\textbf{CPU}}                     & \textbf{Factor} \\ \hline
\textbf{RAKA}                          & Ant colony approach of \cite{Jovanovic2010}       & Not available                                                          & --                       \\ 
\textbf{HGA}                           & Hybrid genetic algorithm of \cite{Potluri2013} & Not available                                                          & --                       \\
\textbf{ACO-LS}                        & Hybrid ACO with local search of  \cite{Potluri2013}  & Not available                                                          & --                       \\ 
\textbf{ACO-PP-LS}                     & Hybrid ACO with pre-processing of \cite{Potluri2013}   & Not available                                                          & --                       \\
\textbf{HMA}                        &  Hybrid memetic algorithm of \cite{Lin2016} &  AMD 3.4GHz & 0.33                     \\
\textbf{R-PBIG}                        &Iterated greedy algorithm of \cite{Bouamama2016}  & Xeon 5670 2.93GHz     & 0.67                    \\ 
\textbf{Hyb-R-PBIG}                    & Hybrid algorithm of \cite{Bouamama2016} & Xeon 5670 2.93GHz    & 0.67                    \\
\textbf{C${\text{C}}^2$FS}                         &  Configuration checking algorithm of \cite{Wang2017a} &  I5-3470 3.2GHz & 0.95                    \\
\textbf{MSRS$\text{L}_{0}$}                       &  Multi-start local-search algorithm  of \cite{Chalupa2018} &  I7-5960X 3.0GHz & 1.23  \\
 \textbf{HTS-DS} & Hybrid tabu search of this paper &  I7-5820K 3.3GHz & 1.00 \\
 \hline
\end{tabular}
}
\end{table}

Tables \ref{tab:MWDSt11}--\ref{tab:MWDSt22} now compare the results of all methods. Each table corresponds to a different instance class (SMPI and LPI) and type (T1 and T2), and each row corresponds to a group of ten instances with identical characteristics.
From left to right, the columns report the characteristics of the instances, the average solution quality and CPU time of previous algorithms, as well as the best and average solution quality, and average CPU time of HTS-DS. The two rightmost columns also report the gap of the best (Gap$_\textsc{B}$) and average (Gap$_\textsc{A}$) solutions of HTS-DS, relative to the best known solutions (BKS) in the literature. Let $z$ be the solution value found by HTS-DS and $z_\textsc{bks}$ be the best known solution value, then the percentage gap is computed as  Gap$(\%) = 100 \times(z-z_{\textsc{bks}})/z_{\textsc{bks}}$.
Finally, the best method is highlighted in boldface for each row, and the last line of the table presents some metrics (time and solution quality) averaged over all instances.\\

\begin{table}[htpb] 
\renewcommand{\arraystretch}{1.2}
\setlength{\tabcolsep}{4pt}
\hspace*{-1.2cm}
\rotatebox{90}{
 \begin{varwidth}{\textheight}
\caption{Type T1, Class SMPI -- Comparison of HTS-DS with recent state-of-the-art algorithms}
 \label{tab:MWDSt11}
 \scalebox{0.75}
 {
\begin{tabular}{| r r | r | r r | r r | r r | r r | r r  | r r | r | r  |r r r r r |} \hline       
 &  & \textbf{RAKA} & \multicolumn{2}{c|}{\textbf{HGA}}   & \multicolumn{2}{c|}{\textbf{ACO-LS}}   & \multicolumn{2}{c|}{\textbf{ACO-PP-LS}} 
  &   \multicolumn{2}{c|}{\textbf{HMA}}
 &   \multicolumn{2}{c|}{\textbf{R-PBIG}}   & \multicolumn{2}{c|}{\textbf{Hyb-R-PBIG}} 
 &   \multicolumn{1}{c|}{\textbf{C${\text{C}}^2$FS}$^{*}$}
 &   \multicolumn{1}{c|}{\textbf{MSRL${\text{S}}_0$}$^{\dagger}$}
 & \multicolumn{5}{c|}{\textbf{HTS-DS}}           \\ \hline
\multicolumn{1}{|c}{\textbf{$|V|$}} & \multicolumn{1}{c|}{\textbf{$|E|$}} & \textbf{Avg} & \textbf{Avg} & \textbf{T(s)} & \textbf{Avg} & \textbf{T(s)} & \textbf{Avg} & \textbf{T(s)} & \textbf{Avg} & \textbf{T(s)} & \textbf{Avg} & \textbf{T(s)} &
\textbf{Avg} & \textbf{T(s)}
& \textbf{Avg} 
& \textbf{Avg}  & \textbf{Best} & \textbf{Avg} & \textbf{T(s)} & \textbf{$\text{Gap}_{\textsc{B}}$} & \textbf{$\text{Gap}_{\textsc{A}}$} \\ \hline

50 & 50 & 539.8 & \textbf{531.3} & 2.7 & \textbf{531.3} & 1.2 & 532.6 & 1.1 & 531.8 & 0.2 & \textbf{531.3} & 0.3 & \textbf{531.3} & 0.3 & \textbf{531.3} & 531.3 & \textbf{531.3} & \textbf{531.3} & 0.1 & 0 & 0 \\ 
 & 100 & 391.9 & 371.2 & 2.6 & 371.2 & 1.0 & 371.5 & 0.9 & 371.2 & 0.2 & 371.1 & 0.5 & \textbf{370.9} & 0.5 & \textbf{370.9} & 370.9 & \textbf{370.9} & \textbf{370.9} & 0.1 & 0 & 0 \\ 
 & 250 & 195.3 & 175.7 & 2.3 & 176.0 & 0.6 & \textbf{175.7} & 0.6 & 176.4 & 0.2 & \textbf{175.7} & 0.9 & \textbf{175.7} & 0.9 & \textbf{175.7} & 175.7 & \textbf{175.7} & \textbf{175.7} & 0.1 & 0 & 0 \\ 
 & 500 & 112.8 & 94.9 & 2.0 & 94.9 & 0.5 & 95.2 & 0.5 & 96.2 & 0.2 & 95.0 & 1.5 & \textbf{94.9} & 1.5 & \textbf{94.9} & 94.9 & \textbf{94.9} & \textbf{94.9} & 0.1 & 0 & 0 \\ 
 & 750 & 69.0 & 63.1 & 2.1 & 63.1 & 0.3 & 63.2 & 0.3 & 63.3 & 0.2 & 63.8 & 1.7 & \textbf{63.1} & 1.7 & 63.3 & 63.1 & \textbf{63.1} & \textbf{63.1} & 0.1 & 0 & 0 \\ 
 & 1000 & 44.7 & 41.5 & 2.1 & 41.5 & 0.3 & \textbf{41.5} & 0.3 & 41.5 & 0.2 & \textbf{41.5} & 2.0 & \textbf{41.5} & 2.0 & \textbf{41.5} & 43.2 & \textbf{41.5} & \textbf{41.5} & 0.0 & 0 & 0 \\ 
100 & 100 & 1087.2 & 1081.3 & 9.6 & 1066.9 & 4.3 & 1065.4 & 3.9 & 1064.9 & 0.5 & 1061.9 & 0.9 & \textbf{1061.0} & 0.9 & \textbf{1061.0} & 1061.0 & \textbf{1061.0} & \textbf{1061.0} & 0.2 & 0 & 0 \\ 
 & 250 & 698.7 & 626.3 & 8.4 & 627.3 & 3.1 & 627.4 & 2.8 & 623.1 & 0.5 & 619.3 & 1.5 & \textbf{618.9} & 1.5 & \textbf{618.9} & 618.9 & \textbf{618.9} & \textbf{618.9} & 0.5 & 0 & 0 \\ 
 & 500 & 442.8 & 358.3 & 5.8 & 362.5 & 2.3 & 363.2 & 2.0 & 356.8 & 0.4 & 356.5 & 2.0 & \textbf{355.6} & 2.0 & \textbf{355.6} & 355.6 & \textbf{355.6} & \textbf{355.6} & 0.6 & 0 & 0 \\ 
 & 750 & 313.7 & 261.2 & 4.8 & 263.5 & 2.0 & 265.0 & 1.9 & 258.4 & 0.4 & 256.5 & 2.5 & \textbf{255.8} & 2.5 & \textbf{255.8} & 255.8 & \textbf{255.8} & \textbf{255.8} & 0.7 & 0 & 0 \\ 
 & 1000 & 247.8 & 205.6 & 4.9 & 209.2 & 1.7 & 208.8 & 1.7 & 205.9 & 0.3 & 203.6 & 2.9 & \textbf{203.6} & 2.9 & \textbf{203.6} & 203.6 & \textbf{203.6} & \textbf{203.6} & 0.9 & 0 & 0 \\ 
 & 2000 & 125.9 & 108.2 & 4.9 & 108.1 & 1.2 & 108.4 & 1.2 & 107.8 & 0.6 & 108.0 & 4.2 & \textbf{107.4} & 4.2 & \textbf{107.4} & 107.4 & \textbf{107.4} & \textbf{107.4} & 0.6 & 0 & 0 \\ 
150 & 150 & 1630.1 & 1607.0 & 23.4 & 1582.8 & 9.9 & 1585.2 & 8.8 & 1585.3 & 1.0 & 1582.5 & 1.6 & \textbf{1580.5} & 1.6 & \textbf{1580.5} & 1582.1 & \textbf{1580.5} & \textbf{1580.5} & 0.5 & 0 & 0 \\ 
 & 250 & 1317.7 & 1238.6 & 20.8 & 1237.2 & 8.6 & 1238.3 & 7.5 & 1231.8 & 0.6 & 1219.5 & 2.2 & \textbf{1218.2} & 2.2 & \textbf{1218.2} & 1221.7 & \textbf{1218.2} & \textbf{1218.2} & 0.6 & 0 & 0 \\ 
 & 500 & 899.9 & 763.0 & 15.3 & 767.7 & 6.1 & 768.6 & 5.5 & 749.5 & 0.5 & 745.0 & 2.9 & \textbf{744.6} & 2.9 & \textbf{744.6} & 746.9 & \textbf{744.6} & \textbf{744.6} & 3.0 & 0 & 0 \\ 
 & 750 & 674.4 & 558.5 & 12.2 & 565.0 & 5.0 & 562.8 & 4.5 & 550.2 & 0.5 & 548.3 & 3.5 & 546.8 & 3.5 & \textbf{546.1} & 549.1 & \textbf{546.1} & \textbf{546.1} & 7.0 & 0 & 0 \\ 
 & 1000 & 540.7 & 438.7 & 11.8 & 446.8 & 4.5 & 448.3 & 4.0 & 435.7 & 0.5 & 433.6 & 4.0 & 433.1 & 4.0 & 432.9 & 434.9 & \textbf{432.8} & \textbf{432.8} & 8.1 & \textbf{-0.023} & \textbf{-0.023} \\ 
 & 2000 & 293.1 & 245.7 & 9.2 & 259.4 & 3.3 & 255.6 & 3.3 & 244.2 & 0.7 & 241.5 & 5.8 & \textbf{241.8} & 5.8 & \textbf{240.8} & 241.1 & \textbf{240.8} & \textbf{240.8} & 7.8 & 0 & 0 \\ 
 & 3000 & 204.7 & 169.2 & 8.3 & 173.4 & 3.0 & 175.2 & 2.8 & 168.4 & 1.3 & 168.4 & 7.4 & 167.8 & 7.4 & \textbf{166.9} & 166.9 & \textbf{166.9} & \textbf{166.9} & 7.5 & 0 & 0 \\ 
200 & 250 & 2039.2 & 1962.1 & 41.7 & 1934.3 & 17.7 & 1927.0 & 15.3 & 1912.1 & 1.0 & 1914.6 & 3.1 & \textbf{1909.7} & 3.1 & 1910.4 & 1917.9 & \textbf{1909.7} & \textbf{1909.7} & 0.6 & 0 & 0 \\ 
 & 500 & 1389.4 & 1266.3 & 33.4 & 1259.7 & 13.2 & 1260.8 & 11.5 & 1245.7 & 0.8 & 1235.3 & 4.2 & 1234.0 & 4.2 & \textbf{1232.8} & 1242.1 & \textbf{1232.8} & \textbf{1232.8} & 2.0 & 0 & 0 \\ 
 & 750 & 1096.2 & 939.8 & 28.1 & 938.7 & 10.6 & 940.1 & 9.2 & 926.1 & 0.7 & 914.9 & 5.0 & 913.8 & 5.0 & \textbf{911.2} & 923.1 & \textbf{911.2} & \textbf{911.2} & 7.8 & 0 & 0 \\ 
 & 1000 & 869.9 & 747.8 & 24.9 & 751.2 & 9.0 & 753.7 & 8.0 & 727.4 & 0.9 & 725.2 & 5.5 & 726.0 & 5.5 & 724.0 & 737.7 & \textbf{723.5} & \textbf{723.5} & 7.8 & \textbf{-0.069} & \textbf{-0.069} \\ 
 & 2000 & 524.1 & 432.9 & 15.2 & 440.2 & 6.5 & 444.7 & 6.0 & 421.2 & 0.9 & 414.8 & 7.3 & 414.7 & 7.3 & \textbf{412.7} & 423.1 & \textbf{412.7} & 412.9 & 8.0 & 0 & 0.048 \\ 
 & 3000 & 385.7 & 308.5 & 14.4 & 309.9 & 5.7 & 315.2 & 5.4 & 297.9 & 1.3 & 294.2 & 9.5 & 296.0 & 9.5 & \textbf{292.8} & 298.4 & \textbf{292.8} & \textbf{292.8} & 7.2 & 0 & 0 \\ 
250 & 250 & -- & 2703.4 & 72.7 & 2655.4 & 32.5 & 2655.4 & 28.4 & 2653.4 & 2.0 & 2653.7 & 4.0 & \textbf{2633.0} & 4.0 & 2633.4 & 2646.7 & \textbf{2633.0} & \textbf{2633.0} & 1.1 & 0 & 0 \\ 
 & 500 & -- & 1878.8 & 59.9 & 1850.3 & 25.1 & 1847.9 & 21.9 & 1828.5 & 1.4 & 1812.6 & 5.8 & 1806.1 & 5.8 & \textbf{1805.9} & 1827.1 & \textbf{1805.9} & \textbf{1805.9} & 2.4 & 0 & 0 \\ 
 & 750 & -- & 1421.1 & 55.0 & 1405.2 & 20.0 & 1405.5 & 17.4 & 1389.4 & 1.1 & 1368.6 & 6.4 & 1366.9 & 6.4 & 1362.2 & 1392.9 & \textbf{1361.9} & \textbf{1361.9} & 8.2 & \textbf{-0.022} & \textbf{-0.022} \\ 
 & 1000 & -- & 1143.4 & 47.3 & 1127.1 & 17.3 & 1122.9 & 15.2 & 1109.5 & 0.9 & 1097.1 & 7.3 & 1092.8 & 7.3 & 1091.1 & 1119.6 & \textbf{1089.9} & 1090.1 & 7.9 & \textbf{-0.110} & \textbf{-0.092} \\ 
 & 2000 & -- & 656.6 & 26.1 & 672.8 & 11.6 & 676.4 & 10.7 & 635.3 & 1.0 & 642.1 & 9.4 & 624.3 & 9.4 & 621.9 & 651.5 & \textbf{621.6} & 621.9 & 8.4 & \textbf{-0.048} & 0 \\ 
 & 3000 & -- & 469.3 & 23.0 & 474.1 & 9.5 & 476.3 & 9.0 & 456.6 & 1.1 & 451.5 & 12.0 & 452.5 & 12.0 & \textbf{447.9} & 466.4 & 448.0 & 448.0 & 7.6 & 0.022 & 0.022 \\ 
 & 5000 & -- & 300.5 & 21.8 & 310.4 & 8.4 & 308.7 & 8.1 & 292.8 & 2.9 & 291.5 & 17.0 & 293.1 & 17.0 & \textbf{289.5} & 297.8 & \textbf{289.5} & 289.6 & 7.8 & 0 & 0.035 \\ \hline
 & Avg & -- & 724.1 & 19.3 & 721.2 & 7.7 & 721.5 & 6.9 & 711.2 & 0.8 & 707.5 & 4.5 & 705.5 & 4.5 & 705.5 & 711.5 & 704.4 & 704.5 & 3.6 & \textbf{-0.008} & \textbf{-0.003} \\ \hline
\multicolumn{20}{l}{* -- C${\text{C}}^2$FS was run until a fixed time limit of 50s (equivalent to 47.5s  after CPU scaling)} \vspace*{-0.15cm} \\
\multicolumn{20}{l}{$\dagger$ -- MSRL${\text{S}}_0$ was run until a fixed time limit of 3600s (equivalent to 4428s after CPU scaling)} \\
   \end{tabular}     
}    
 \end{varwidth}
}
 \end{table}

\begin{table}[htpb] 
\renewcommand{\arraystretch}{1.2}
\setlength{\tabcolsep}{3pt}
\hspace*{1.3cm}
\rotatebox{90}{
 \begin{varwidth}{\textheight}
\caption{Type T1, Class LPI -- Comparison of HTS-DS with recent state-of-the-art algorithms}
 \label{tab:MWDSt21}
 \scalebox{0.74}
 {
\begin{tabular}{| r r | r | r r | r r | r r | r r | r r  | r r | r | r  |r r r r r |} \hline       
 &  & \textbf{RAKA} & \multicolumn{2}{c|}{\textbf{HGA}}   & \multicolumn{2}{c|}{\textbf{ACO-LS}}   & \multicolumn{2}{c|}{\textbf{ACO-PP-LS}} 
  &   \multicolumn{2}{c|}{\textbf{HMA}}
 &   \multicolumn{2}{c|}{\textbf{R-PBIG}}   & \multicolumn{2}{c|}{\textbf{Hyb-R-PBIG}} 
 &   \multicolumn{1}{c|}{\textbf{C${\text{C}}^2$FS}$^{*}$}
 &   \multicolumn{1}{c|}{\textbf{MSRL${\text{S}}_0$}$^{\dagger}$}
 & \multicolumn{5}{c|}{\textbf{HTS-DS}}           \\ \hline
\multicolumn{1}{|c}{\textbf{$|V|$}} & \multicolumn{1}{c|}{\textbf{$|E|$}} & \textbf{Avg} & \textbf{Avg} & \textbf{T(s)} & \textbf{Avg} & \textbf{T(s)} & \textbf{Avg} & \textbf{T(s)} & \textbf{Avg} & \textbf{T(s)} & \textbf{Avg} & \textbf{T(s)} &
\textbf{Avg} & \textbf{T(s)}
& \textbf{Avg} 
& \textbf{Avg}  & \textbf{Best} & \textbf{Avg} & \textbf{T(s)} & \textbf{$\text{Gap}_{\textsc{B}}$} & \textbf{$\text{Gap}_{\textsc{A}}$} \\ \hline

300 & 300 & -- & 3255.2 & 116.3 & 3198.5 & 49.2 & 3205.9 & 42.2 & 3199.3 & 2.6 & 3189.3 & 5.1 & \textbf{3175.4} & 5.1 & 3178.6 & 3202.4 & \textbf{3175.4} & \textbf{3175.4} & 1.3 & 0 & 0 \\ 
 & 500 & -- & 2509.8 & 109.0 & 2479.2 & 41.0 & 2473.3 & 35.7 & 2464.4 & 2.9 & 2446.9 & 6.8 & \textbf{2435.6} & 6.8 & 2438.1 & 2468.6 & \textbf{2435.6} & \textbf{2435.6} & 1.6 & 0 & 0 \\ 
 & 750 & -- & 1933.9 & 93.2 & 1903.3 & 34.0 & 1913.9 & 30.0 & 1884.6 & 1.6 & 1869.6 & 8.0 & 1856.8 & 8.0 & 1854.6 & 1897.1 & \textbf{1853.8} & 1853.9 & 7.5 & \textbf{-0.043} & \textbf{-0.038} \\ 
 & 1000 & -- & 1560.1 & 80.1 & 1552.5 & 28.2 & 1555.8 & 24.5 & 1518.4 & 1.6 & 1503.4 & 8.8 & 1498.6 & 8.8 & 1495.0 & 1539.2 & \textbf{1494.0} & 1494.1 & 8.7 & \textbf{-0.067} & \textbf{-0.060} \\ 
 & 2000 & -- & 909.6 & 47.6 & 916.8 & 18.8 & 916.5 & 17.0 & 878.7 & 1.5 & 872.5 & 11.3 & 870.1 & 11.3 & 862.5 & 901.3 & \textbf{862.4} & 862.4 & 8.9 & \textbf{-0.012} & \textbf{-0.012} \\ 
 & 3000 & -- & 654.9 & 37.3 & 667.8 & 15.4 & 670.7 & 14.3 & 640.9 & 1.4 & 629.0 & 14.3 & 628.5 & 14.3 & 624.3 & 663.5 & \textbf{624.1} & 624.4 & 9.5 & \textbf{-0.032} & 0.016 \\ 
 & 5000 & -- & 428.3 & 27.4 & 437.4 & 12.5 & 435.9 & 12.0 & 411.7 & 2.3 & 409.4 & 20.0 & 410.0 & 20.0 & \textbf{406.1} & 428.5 & \textbf{406.1} & 406.3 & 8.1 & 0 & 0.049 \\ 
500 & 500 & 5476.3 & 5498.3 & 412.3 & 5398.3 & 180.3 & 5387.7 & 156.0 & 5392.1 & 6.5 & 5378.4 & 14.1 & \textbf{5304.7} & 14.1 & 5305.7 & 5389.2 & \textbf{5304.7} & \textbf{5304.7} & 2.7 & 0 & 0 \\ 
 & 1000 & 4069.8 & 3798.6 & 359.8 & 3714.8 & 143.7 & 3698.3 & 116.4 & 3678.3 & 4.5 & 3642.2 & 19.5 & \textbf{3607.3} & 19.5 & 3607.8 & 3716.3 & 3607.6 & 3608.6 & 10.4 & 0.008 & 0.036 \\ 
 & 2000 & 2627.5 & 2338.2 & 219.2 & 2277.6 & 90.3 & 2275.9 & 81.7 & 2223.7 & 3.6 & 2203.9 & 24.2 & 2197.2 & 24.2 & 2181.0 & 2291.2 & \textbf{2176.8} & 2177.8 & 10.8 & \textbf{-0.193} & \textbf{-0.147} \\ 
 & 5000 & 1398.5 & 1122.7 & 114.4 & 1115.3 & 51.3 & 1110.2 & 45.7 & 1074.2 & 4.2 & 1055.9 & 37.5 & 1052.1 & 37.5 & 1043.3 & 1138.1 & \textbf{1042.3} & 1044.2 & 10.9 & \textbf{-0.096} & 0.086 \\ 
 & 10000 & 825.7 & 641.1 & 64.0 & 652.8 & 36.9 & 650.9 & 35.8 & 595.4 & 9.4 & 596.3 & 51.1 & 597.5 & 51.1 & \textbf{587.2} & 634.5 & \textbf{587.2} & \textbf{587.2} & 9.9 & 0 & 0 \\ 
800 & 1000 & 8098.9 & 8017.7 & 1459.5 & 8117.6 & 769.8 & 8068.0 & 709.6 & 7839.9 & 15.4 & 7768.6 & 45.5 & \textbf{7655.0} & 45.5 & 7663.4 & 7833.7 & \textbf{7655.0} & \textbf{7655.0} & 5.2 & 0 & 0 \\ 
 & 2000 & 5739.9 & 5318.7 & 1094.3 & 5389.9 & 572.6 & 5389.6 & 554.8 & 5100.7 & 8.6 & 5037.9 & 55.8 & 5002.8 & 55.8 & \textbf{4982.1} & 5224.6 & 4987.3 & 4991.7 & 14.3 & 0.104 & 0.193 \\ 
 & 5000 & 3116.5 & 2633.4 & 551.5 & 2616.0 & 263.5 & 2607.9 & 237.3 & 2495.7 & 9.9 & 2465.4 & 82.1 & 2469.2 & 82.1 & 2441.2 & 2626.9 & \textbf{2432.6} & 2435.8 & 14.2 & \textbf{-0.352} & \textbf{-0.221} \\ 
 & 10000 & 1923.0 & 1547.7 & 246.1 & 1525.7 & 147.5 & 1535.3 & 145.4 & 1459.8 & 9.9 & 1420.0 & 114.6 & 1414.8 & 114.6 & 1395.6 & 1526.2 & \textbf{1393.7} & 1395.1 & 13.9 & \textbf{-0.136} & \textbf{-0.036} \\ 
1000 & 1000 & 10924.4 & 11095.2 & 2829.6 & 11035.5 & 1320.7 & 11022.9 & 1189.9 & 10863.3 & 27.6 & 10825.2 & 64.9 & \textbf{10574.4} & 64.9 & 10585.3 & 10766.7 & \textbf{10574.4} & \textbf{10574.4} & 8.7 & 0 & 0 \\ 
 & 5000 & 4662.7 & 3996.6 & 1152.1 & 4012.0 & 627.7 & 4029.8 & 600.8 & 3742.8 & 18.2 & 3693.1 & 123.4 & 3699.7 & 123.4 & 3671.8 & 3947.0 & \textbf{3656.6} & 3662.7 & 15.7 & \textbf{-0.414} & \textbf{-0.248} \\ 
 & 10000 & 2890.3 & 2334.7 & 566.2 & 2314.9 & 308.8 & 2306.6 & 289.9 & 2193.7 & 18.0 & 2140.3 & 170.8 & 2138.1 & 170.8 & 2109.0 & 2283.2 & \textbf{2099.8} & 2102.2 & 15.9 & \textbf{-0.436} & \textbf{-0.322} \\ 
 & 15000 & 2164.3 & 1687.5 & 356.3 & 1656.3 & 227.7 & 1657.4 & 211.0 & 1590.9 & 28.2 & 1549.1 & 188.9 & 1548.0 & 188.9 & 1521.5 & -- & \textbf{1519.7} & 1521.9 & 16.2 & \textbf{-0.118} & 0.026 \\ 
 & 20000 & 1734.3 & 1337.2 & 251.1 & 1312.8 & 204.9 & 1315.8 & 200.3 & 1263.5 & 32.6 & 1219.0 & 194.2 & 1216.9 & 194.2 & 1203.6 & -- & \textbf{1200.9} & 1205.6 & 17.9 & \textbf{-0.224} & 0.166 \\ \hline
 & Avg & -- & 2981.9 & 485.1 & 2966.4 & 245.0 & 2963.3 & 226.2 & 2881.5 & 10.0 & 2853.1 & 60.0 & 2826.3 & 60.0 & 2817.0 & -- & 2813.8 & 2815.2 & 10.1 & \textbf{-0.096} & \textbf{-0.024} \\ \hline
\multicolumn{20}{l}{* -- C${\text{C}}^2$FS was run until a fixed time limit of 50s when $|V| \leq 500$, and 1000s otherwise, (equivalent to 47.5s and 950s after CPU scaling, respectively)} \vspace*{-0.15cm} \\
\multicolumn{20}{l}{$\dagger$ -- MSRL${\text{S}}_0$ was run until a fixed time limit of 3600s (equivalent to 4428s after CPU scaling)} \\
\end{tabular}
}
 \end{varwidth}
}
\end{table}

\begin{table}[htpb] 
\renewcommand{\arraystretch}{1.2}
\setlength{\tabcolsep}{4.0pt}
\hspace*{-1cm}
\rotatebox{90}{
 \begin{varwidth}{\textheight}
\caption{Type T2, Class SMPI -- Comparison of HTS-DS with recent state-of-the-art algorithms}
 \label{tab:MWDSt12} 
 \scalebox{0.75}
 {
 \begin{tabular}{| r r | r | r r | r r | r r | r r | r r  | r r | r | r  |r r r r r |} \hline       
 &  & \textbf{RAKA} & \multicolumn{2}{c|}{\textbf{HGA}}   & \multicolumn{2}{c|}{\textbf{ACO-LS}}   & \multicolumn{2}{c|}{\textbf{ACO-PP-LS}} 
  &   \multicolumn{2}{c|}{\textbf{HMA}}
 &   \multicolumn{2}{c|}{\textbf{R-PBIG}}   & \multicolumn{2}{c|}{\textbf{Hyb-R-PBIG}} 
 &   \multicolumn{1}{c|}{\textbf{C${\text{C}}^2$FS}$^{*}$}
 &   \multicolumn{1}{c|}{\textbf{MSRL${\text{S}}_0$}$^{\dagger}$}
 & \multicolumn{5}{c|}{\textbf{HTS-DS}}           \\ \hline
\multicolumn{1}{|c}{\textbf{$|V|$}} & \multicolumn{1}{c|}{\textbf{$|E|$}} & \textbf{Avg} & \textbf{Avg} & \textbf{T(s)} & \textbf{Avg} & \textbf{T(s)} & \textbf{Avg} & \textbf{T(s)} & \textbf{Avg} & \textbf{T(s)} & \textbf{Avg} & \textbf{T(s)} &
\textbf{Avg} & \textbf{T(s)}
& \textbf{Avg} 
& \textbf{Avg}  & \textbf{Best} & \textbf{Avg} & \textbf{T(s)} & \textbf{$\text{Gap}_{\textsc{B}}$} & \textbf{$\text{Gap}_{\textsc{A}}$} \\ \hline

50 & 50 & 62.3 & \textbf{60.8} & 3.8 & \textbf{60.8} & 1.2 & \textbf{60.8} & 1.2 & 60.8 & 0.3 & \textbf{60.8} & 0.3 & \textbf{60.8} & 0.3 & \textbf{60.8} & 60.8 & \textbf{60.8} & \textbf{60.8} & 0.1 & 0 & 0 \\ 
 & 100 & 98.4 & \textbf{90.3} & 3.8 & \textbf{90.3} & 1.1 & \textbf{90.3} & 1.1 & 90.3 & 0.3 & \textbf{90.3} & 0.5 & \textbf{90.3} & 0.5 & \textbf{90.3} & 90.3 & \textbf{90.3} & \textbf{90.3} & 0.1 & 0 & 0 \\ 
 & 250 & 202.4 & \textbf{146.7} & 3.4 & \textbf{146.7} & 0.8 & \textbf{146.7} & 0.7 & 146.7 & 0.2 & \textbf{146.7} & 0.9 & \textbf{146.7} & 0.9 & \textbf{146.7} & 146.7 & \textbf{146.7} & \textbf{146.7} & 0.0 & 0 & 0 \\ 
 & 500 & 312.9 & \textbf{179.9} & 2.8 & \textbf{179.9} & 0.6 & \textbf{179.9} & 0.6 & 179.9 & 0.2 & \textbf{179.9} & 1.4 & \textbf{179.9} & 1.4 & \textbf{179.9} & 179.9 & \textbf{179.9} & \textbf{179.9} & 0.0 & 0 & 0 \\ 
 & 750 & 386.3 & \textbf{171.1} & 2.5 & \textbf{171.1} & 0.5 & \textbf{171.1} & 0.4 & 171.1 & 0.2 & \textbf{171.1} & 1.6 & \textbf{171.1} & 1.6 & \textbf{171.1} & 171.1 & \textbf{171.1} & \textbf{171.1} & 0.0 & 0 & 0 \\ 
 & 1000 & -- & \textbf{146.5} & 1.9 & \textbf{146.5} & 0.3 & \textbf{146.5} & 0.2 & 146.5 & 0.1 & \textbf{146.5} & 1.9 & \textbf{146.5} & 1.9 & \textbf{146.5} & 146.5 & \textbf{146.5} & \textbf{146.5} & 0.1 & 0 & 0 \\ 
100 & 100 & 126.5 & 124.5 & 11.0 & 123.6 & 4.2 & \textbf{123.5} & 4.0 & 124.4 & 0.8 & \textbf{123.5} & 0.8 & \textbf{123.5} & 0.8 & \textbf{123.5} & 123.5 & \textbf{123.5} & \textbf{123.5} & 0.2 & 0 & 0 \\ 
 & 250 & 236.6 & 211.4 & 11.0 & 210.2 & 3.5 & 210.4 & 3.2 & 210.4 & 0.7 & \textbf{209.2} & 1.4 & \textbf{209.2} & 1.4 & \textbf{209.2} & 210.8 & \textbf{209.2} & \textbf{209.2} & 0.1 & 0 & 0 \\ 
 & 500 & 404.8 & 306.0 & 10.2 & 307.8 & 2.7 & 308.4 & 2.5 & 307.1 & 0.5 & \textbf{305.7} & 1.9 & \textbf{305.7} & 1.9 & \textbf{305.7} & 305.9 & \textbf{305.7} & \textbf{305.7} & 0.1 & 0 & 0 \\ 
 & 750 & 615.1 & 385.3 & 8.6 & 385.7 & 2.4 & 386.3 & 2.3 & 384.8 & 0.4 & 386.9 & 2.3 & \textbf{384.5} & 2.3 & \textbf{384.5} & 384.5 & \textbf{384.5} & \textbf{384.5} & 0.2 & 0 & 0 \\ 
 & 1000 & 697.3 & 429.1 & 8.5 & 430.3 & 2.0 & 430.3 & 1.9 & 428.0 & 0.4 & \textbf{427.3} & 2.8 & \textbf{427.3} & 2.8 & \textbf{427.3} & 427.3 & \textbf{427.3} & \textbf{427.3} & 0.2 & 0 & 0 \\ 
 & 2000 & 1193.9 & 550.6 & 7.8 & 558.8 & 1.6 & 559.8 & 1.5 & 552.2 & 0.3 & 552.7 & 4.2 & \textbf{550.6} & 4.2 & \textbf{550.6} & 551.0 & \textbf{550.6} & \textbf{550.6} & 0.3 & 0 & 0 \\ 
150 & 150 & 190.1 & 186.0 & 29.5 & 184.7 & 9.1 & 184.9 & 8.6 & 185.8 & 1.7 & \textbf{184.5} & 1.4 & \textbf{184.5} & 1.4 & \textbf{184.5} & 184.5 & \textbf{184.5} & \textbf{184.5} & 0.5 & 0 & 0 \\ 
 & 250 & 253.9 & 234.9 & 29.1 & 233.2 & 8.6 & 233.4 & 7.8 & 234.1 & 1.6 & \textbf{232.8} & 2.1 & \textbf{232.8} & 2.1 & \textbf{232.8} & 234.0 & \textbf{232.8} & \textbf{232.8} & 0.3 & 0 & 0 \\ 
 & 500 & 443.2 & 350.0 & 24.5 & 351.9 & 7.0 & 351.9 & 6.2 & 350.9 & 1.0 & 349.7 & 2.9 & \textbf{349.5} & 2.9 & \textbf{349.5} & 353.9 & \textbf{349.5} & \textbf{349.5} & 0.3 & 0 & 0 \\ 
 & 750 & 623.3 & 455.8 & 21.8 & 456.9 & 5.9 & 454.7 & 5.6 & 453.5 & 0.8 & \textbf{452.4} & 3.6 & \textbf{452.4} & 3.6 & \textbf{452.4} & 458.9 & \textbf{452.4} & \textbf{452.4} & 0.4 & 0 & 0 \\ 
 & 1000 & 825.3 & 547.5 & 19.7 & 551.4 & 5.5 & 549.0 & 5.1 & 549.4 & 0.8 & 547.8 & 4.0 & \textbf{547.2} & 4.0 & \textbf{547.2} & 549.1 & \textbf{547.2} & \textbf{547.2} & 0.4 & 0 & 0 \\ 
 & 2000 & 1436.4 & 720.1 & 17.5 & 725.7 & 4.4 & 725.7 & 4.1 & 722.2 & 0.8 & \textbf{720.1} & 5.6 & \textbf{720.1} & 5.6 & \textbf{720.1} & 720.1 & \textbf{720.1} & \textbf{720.1} & 0.5 & 0 & 0 \\ 
 & 3000 & 1751.9 & 792.6 & 15.4 & 794.0 & 3.6 & 806.2 & 3.6 & 797.8 & 0.5 & 793.2 & 7.8 & \textbf{792.4} & 7.8 & \textbf{792.4} & 792.6 & \textbf{792.4} & \textbf{792.4} & 0.8 & 0 & 0 \\ 
200 & 250 & 293.2 & 275.1 & 53.9 & 272.6 & 17.1 & 272.6 & 15.6 & 273.8 & 2.0 & \textbf{271.7} & 2.9 & \textbf{271.7} & 2.9 & \textbf{271.7} & 273.2 & \textbf{271.7} & \textbf{271.7} & 0.5 & 0 & 0 \\ 
 & 500 & 456.5 & 390.7 & 49.9 & 388.6 & 14.6 & 388.4 & 12.8 & 389.1 & 1.7 & 386.8 & 4.1 & \textbf{386.7} & 4.1 & \textbf{386.7} & 401.8 & \textbf{386.7} & \textbf{386.7} & 0.4 & 0 & 0 \\ 
 & 750 & 657.9 & 507.0 & 45.9 & 501.7 & 12.5 & 501.4 & 11.1 & 500.3 & 1.5 & 497.1 & 4.8 & \textbf{497.1} & 4.8 & \textbf{497.1} & 503.4 & \textbf{497.1} & \textbf{497.1} & 0.4 & 0 & 0 \\ 
 & 1000 & 829.2 & 601.1 & 37.3 & 605.9 & 11.2 & 605.8 & 10.2 & 606.6 & 1.1 & 596.8 & 5.7 & \textbf{596.8} & 5.7 & \textbf{596.8} & 609.7 & \textbf{596.8} & \textbf{596.8} & 0.5 & 0 & 0 \\ 
 & 2000 & 1626.0 & 893.5 & 31.9 & 891.0 & 8.9 & 892.9 & 8.1 & 890.3 & 0.6 & 884.6 & 7.6 & \textbf{884.6} & 7.6 & \textbf{884.6} & 890.2 & \textbf{884.6} & \textbf{884.6} & 0.6 & 0 & 0 \\ 
 & 3000 & 2210.3 & 1021.3 & 29.9 & 1027.0 & 7.2 & 1034.4 & 7.0 & 1026.2 & 0.5 & 1019.2 & 9.4 & \textbf{1019.2} & 9.4 & \textbf{1019.2} & 1029.2 & \textbf{1019.2} & \textbf{1019.2} & 1.0 & 0 & 0 \\ 
250 & 250 & -- & 310.1 & 88.0 & 306.5 & 28.0 & 306.7 & 25.5 & 310.4 & 3.3 & \textbf{306} & 3.3 & \textbf{306.0} & 3.3 & 306.1 & 306.6 & \textbf{306.0} & \textbf{306.0} & 1.1 & 0 & 0 \\ 
 & 500 & -- & 444.0 & 89.1 & 443.8 & 25.4 & 443.2 & 22.8 & 445.8 & 1.8 & 441.0 & 5.5 & \textbf{440.7} & 5.5 & \textbf{440.7} & 454.3 & \textbf{440.7} & \textbf{440.7} & 0.7 & 0 & 0 \\ 
 & 750 & -- & 578.2 & 77.1 & 573.1 & 23.0 & 575.9 & 20.6 & 573.1 & 1.8 & 567.9 & 6.7 & \textbf{567.4} & 6.7 & \textbf{567.4} & 583.8 & \textbf{567.4} & \textbf{567.4} & 0.7 & 0 & 0 \\ 
 & 1000 & -- & 672.8 & 74.2 & 671.8 & 20.8 & 675.1 & 19.0 & 676.7 & 1.4 & 669.2 & 7.6 & \textbf{668.6} & 7.6 & \textbf{668.6} & 689.6 & \textbf{668.6} & \textbf{668.6} & 0.7 & 0 & 0 \\ 
 & 2000 & -- & 1030.8 & 56.8 & 1033.9 & 15.9 & 1031.5 & 15.1 & 1025.8 & 1.1 & 1009.5 & 9.7 & \textbf{1007.0} & 9.7 & \textbf{1007.0} & 1044.7 & \textbf{1007.0} & \textbf{1007.0} & 1.1 & 0 & 0 \\ 
 & 3000 & -- & 1262.0 & 49.0 & 1288.5 & 14.0 & 1277.0 & 13.5 & 1261.4 & 0.7 & 1251.6 & 12.1 & \textbf{1250.6} & 12.1 & \textbf{1250.6} & 1262.4 & \textbf{1250.6} & \textbf{1250.6} & 1.4 & 0 & 0 \\ 
 & 5000 & -- & 1480.9 & 46.9 & 1493.6 & 11.2 & 1520.1 & 10.9 & 1484.0 & 0.8 & \textbf{1464.2} & 17.1 & \textbf{1464.2} & 17.1 & \textbf{1464.2} & 1472.9 & \textbf{1464.2} & \textbf{1464.2} & 2.2 & 0 & 0 \\ \hline
 & Avg & -- & 486.1 & 30.1 & 487.7 & 8.6 & 488.9 & 7.9 & 486.2 & 0.9 & 482.7 & 4.5 & 482.4 & 4.5 & 482.4 & 487.9 & 482.4 & 482.4 & 0.5 & 0 & 0 \\ \hline
\multicolumn{20}{l}{* -- C${\text{C}}^2$FS was run until a fixed time limit of 50s (equivalent to 47.5s  after CPU scaling)} \vspace*{-0.15cm} \\
\multicolumn{20}{l}{$\dagger$ -- MSRL${\text{S}}_0$ was run until a fixed time limit of 3600s (equivalent to 4428s after CPU scaling)} \\
\end{tabular}     
}              
 \end{varwidth}
}
 \end{table}

\begin{table}[htpb] 
\renewcommand{\arraystretch}{1.2}
\setlength{\tabcolsep}{3.5pt}
\hspace*{1cm}
\rotatebox{90}{
 \begin{varwidth}{\textheight}
\caption{Type T2, Class LPI -- Comparison of HTS-DS with recent state-of-the-art algorithms}
 \label{tab:MWDSt22}
 \scalebox{0.75}
 {
\begin{tabular}{| r r | r | r r | r r | r r | r r | r r  | r r | r | r  |r r r r r |} \hline       
 &  & \textbf{RAKA} & \multicolumn{2}{c|}{\textbf{HGA}}   & \multicolumn{2}{c|}{\textbf{ACO-LS}}   & \multicolumn{2}{c|}{\textbf{ACO-PP-LS}} 
  &   \multicolumn{2}{c|}{\textbf{HMA}}
 &   \multicolumn{2}{c|}{\textbf{R-PBIG}}   & \multicolumn{2}{c|}{\textbf{Hyb-R-PBIG}} 
 &   \multicolumn{1}{c|}{\textbf{C${\text{C}}^2$FS}$^{*}$}
 &   \multicolumn{1}{c|}{\textbf{MSRL${\text{S}}_0$}$^{\dagger}$}
 & \multicolumn{5}{c|}{\textbf{HTS-DS}}           \\ \hline
\multicolumn{1}{|c}{\textbf{$|V|$}} & \multicolumn{1}{c|}{\textbf{$|E|$}} & \textbf{Avg} & \textbf{Avg} & \textbf{T(s)} & \textbf{Avg} & \textbf{T(s)} & \textbf{Avg} & \textbf{T(s)} & \textbf{Avg} & \textbf{T(s)} & \textbf{Avg} & \textbf{T(s)} &
\textbf{Avg} & \textbf{T(s)}
& \textbf{Avg} 
& \textbf{Avg}  & \textbf{Best} & \textbf{Avg} & \textbf{T(s)} & \textbf{$\text{Gap}_{\textsc{B}}$} & \textbf{$\text{Gap}_{\textsc{A}}$} \\ \hline

300 & 300 & -- & 375.6 & 142.3 & 371.1 & 43.0 & 371.1 & 39.2 & 373.9 & 4.7 & \textbf{369.9} & 4.2 & \textbf{369.9} & 4.2 & \textbf{369.9} & 371.2 & \textbf{369.9} & \textbf{369.9} & 1.3 & 0 & 0 \\ 
 & 500 & -- & 484.2 & 143.4 & 480.8 & 40.0 & 481.2 & 35.8 & 484.0 & 3.8 & 478.0 & 6.4 & \textbf{477.8} & 6.4 & \textbf{477.8} & 485.8 & \textbf{477.8} & \textbf{477.8} & 0.9 & 0 & 0 \\ 
 & 750 & -- & 623.8 & 124.5 & 621.6 & 36.9 & 618.3 & 32.9 & 620.1 & 2.4 & 613.6 & 7.8 & \textbf{613.3} & 7.8 & \textbf{613.3} & 634.9 & \textbf{613.3} & \textbf{613.3} & 0.9 & 0 & 0 \\ 
 & 1000 & -- & 751.1 & 113.0 & 744.9 & 33.7 & 743.5 & 30.5 & 745.1 & 1.8 & 738.3 & 9.0 & \textbf{737.7} & 9.0 & 737.9 & 755.8 & \textbf{737.7} & \textbf{737.7} & 1.0 & 0 & 0 \\ 
 & 2000 & -- & 1106.7 & 87.7 & 1111.6 & 24.3 & 1107.5 & 22.7 & 1103.6 & 1.0 & 1094.6 & 11.7 & \textbf{1093.8} & 11.7 & \textbf{1093.8} & 1118.0 & \textbf{1093.8} & \textbf{1093.8} & 1.1 & 0 & 0 \\ 
 & 3000 & -- & 1382.1 & 70.9 & 1422.8 & 23.9 & 1415.3 & 20.9 & 1378.9 & 1.0 & 1358.5 & 14.0 & \textbf{1358.5} & 14.0 & \textbf{1358.5} & 1392.5 & \textbf{1358.5} & \textbf{1358.5} & 1.5 & 0 & 0 \\ 
 & 5000 & -- & 1686.3 & 66.7 & 1712.1 & 18.2 & 1698.6 & 17.4 & 1692.6 & 0.7 & 1683.2 & 19.8 & \textbf{1682.7} & 19.8 & \textbf{1682.7} & 1701.3 & \textbf{1682.7} & \textbf{1682.7} & 2.8 & 0 & 0 \\ 
500 & 500 & 651.2 & 632.9 & 522.0 & 627.5 & 149.1 & 627.3 & 135.4 & 633.4 & 9.3 & 624.2 & 11.9 & \textbf{623.6} & 11.9 & \textbf{623.6} & 627.0 & \textbf{623.6} & \textbf{623.6} & 2.8 & 0 & 0 \\ 
 & 1000 & 1018.1 & 919.2 & 478.8 & 913.0 & 137.8 & 912.6 & 122.1 & 912.0 & 6.8 & 901.3 & 18.8 & \textbf{899.6} & 18.8 & 899.8 & 934.9 & \textbf{899.6} & \textbf{899.6} & 2.0 & 0 & 0 \\ 
 & 2000 & 1871.8 & 1398.2 & 354.3 & 1384.9 & 109.9 & 1383.9 & 98.7 & 1394.1 & 4.2 & 1364.4 & 24.9 & \textbf{1362.2} & 24.9 & 1363.3 & 1408.6 & \textbf{1362.2} & \textbf{1362.2} & 2.4 & 0 & 0 \\ 
 & 5000 & 4299.8 & 2393.2 & 217.5 & 2459.1 & 91.3 & 2468.8 & 90.7 & 2388.3 & 1.6 & 2341.5 & 39.5 & \textbf{2326.6} & 39.5 & \textbf{2333.7} & 2401.7 & \textbf{2326.6} & \textbf{2326.6} & 3.4 & 0 & 0 \\ 
 & 10000 & 8543.5 & 3264.9 & 191.1 & 3377.9 & 60.1 & 3369.4 & 60.4 & 3259.6 & 1.5 & 3216.1 & 53.9 & \textbf{3211.5} & 53.9 & \textbf{3211.5} & 3261.5 & \textbf{3211.5} & \textbf{3211.5} & 8.5 & 0 & 0 \\ 
800 & 1000 & 1171.2 & 1128.2 & 1810.5 & 1126.4 & 498.8 & 1125.1 & 444.9 & 1131.3 & 18.5 & 1107.6 & 39.9 & \textbf{1103.9} & 39.9 & 1104.3 & 1126.2 & \textbf{1103.9} & \textbf{1103.9} & 4.3 & 0 & 0 \\ 
 & 2000 & 1938.7 & 1679.2 & 1560.5 & 1693.7 & 516.6 & 1697.9 & 474.7 & 1681.9 & 10.9 & 1634.6 & 55.9 & \textbf{1630.8} & 55.9 & 1632.3 & 1709.9 & \textbf{1630.8} & \textbf{1630.8} & 4.1 & 0 & 0 \\ 
 & 5000 & 4439.0 & 3003.6 & 955.7 & 3121.9 & 413.7 & 3120.9 & 407.5 & 2963.8 & 6.0 & 2884.8 & 86.0 & \textbf{2876.1} & 86.0 & 2878.5 & 2990.4 & \textbf{2876.1} & \textbf{2876.6} & 10.8 & 0 & 0.017 \\ 
 & 10000 & 8951.1 & 4268.1 & 653.6 & 4404.1 & 249.9 & 4447.9 & 249.7 & 4226.6 & 3.3 & 4103.7 & 123.2 & 4103.3 & 123.2 & 4105.6 & 4272.5 & \textbf{4102.8} & 4104.0 & 21.5 & \textbf{-0.012} & 0.017 \\ 
1000 & 1000 & 1289.3 & 1265.2 & 3481.4 & 1259.3 & 931.2 & 1258.6 & 832.8 & 1270.9 & 25.6 & 1243.6 & 54.2 & \textbf{1237.5} & 54.2 & 1237.7 & 1247.6 & \textbf{1237.5} & \textbf{1237.5} & 8.8 & 0 & 0 \\ 
 & 5000 & 4720.1 & 3320.1 & 1995.4 & 3411.6 & 832.6 & 3415.1 & 827.4 & 3317.6 & 8.2 & 3195.7 & 131.3 & \textbf{3172.9} & 131.3 & 3178.7 & 3340.5 & 3178.2 & 3180.0 & 16.6 & 0.167 & 0.224 \\ 
 & 10000 & 9407.7 & 4947.5 & 1250.5 & 5129.1 & 546.5 & 5101.9 & 548.6 & 4937.9 & 6.9 & 4722.4 & 184.0 & \textbf{4704.8} & 184.0 & 4711.8 & 4935.5 & 4707.2 & 4709.9 & 31.2 & 0.051 & 0.108 \\ 
 & 15000 & 14433.5 & 6267.6 & 977.7 & 6454.6 & 398.7 & 6470.6 & 398.9 & 6122.5 & 6.0 & 5884.2 & 204.5 & \textbf{5856.4} & 204.5 & 5874.2 & -- & 5859.8 & 5862.4 & 36.5 & 0.058 & 0.102 \\ 
 & 20000 & 19172.6 & 7088.5 & 817.3 & 7297.4 & 322.8 & 7340.8 & 325.5 & 6842.9 & 5.0 & 6678.0 & 213.9 & 6655.9 & 213.9 & 6662.1 & -- & \textbf{6655.1} & 6657.9 & 55.6 & \textbf{-0.012} & 0.03 \\ \hline
 & Avg & -- & 2285.1 & 762.6 & 2339.3 & 260.9 & 2341.7 & 248.4 & 2261.0 & 6.2 & 2201.8 & 62.6 & 2195.2 & 62.6 & 2197.7 & -- & 2195.6 & 2196.2 & 10.4 & 0.012 & 0.024 \\ \hline
\multicolumn{20}{l}{* -- C${\text{C}}^2$FS was run until a fixed time limit of 50s when $|V| \leq 500$, and 1000s otherwise, (equivalent to 47.5s and 950s after CPU scaling, respectively)} \vspace*{-0.15cm} \\
\multicolumn{20}{l}{$\dagger$ -- MSRL${\text{S}}_0$ was run until a fixed time limit of 3600s (equivalent to 4428s after CPU scaling)} \\
   \end{tabular}
}
 \end{varwidth}
}
\end{table}
 
For the benchmark instances of type T1 and class SMPI, the HTS-DS identifies all known optimal solution values (known for 17 instances in total), and even finds new best known solutions for $16\%$ of the instances. The algorithm produces solutions of consistently high quality, with an average gap from the previous BKS of $-0.003\%$, meaning that the average solution quality of HTS-DS is better than the best solutions ever found in all prior studies in the literature (i.e., the best solutions found by multiple algorithms, runs, and parameter settings). In a similar fashion, for the class LPI, the HTS-DS algorithm retrieves all known optimal solutions (50 in total), and produces new best solutions for $52\%$ of the instances, with an average percentage gap of  $-0.096\%$ relative to the BKS from previous literature. HTS-DS is also faster than existing algorithms, with an average CPU time of 3.6 and 10.1 seconds on the classes SMPI and LPI, respectively.

Similar observations are valid for the benchmark instances of type T2. For this type, the instances of class SMPI are easier to solve, and most recent methods find the same solutions. In contrast, the class LPI allows to observe significant differences between methods. Again, for this benchmark, HTS-DS retrieves very accurate solutions, with an average gap of $0.024\%$ relative to the BKS, in a time which is significantly smaller than previous approaches.

Finally, the BHOSLIB and DIMACS benchmark instances were extended in \cite{Wang2017a} with a weight $w(i) = (i \bmod 200) + 1$ for each vertex $i$, and the authors provided experimental results of ACO-PP-LS and C${\text{C}}^2$FS on these test sets. We therefore tested HTS-DS on these instances and report the solutions in the appendix of this paper. These results highlight again the excellent performance of the proposed algorithm, which finds or improves all known BKS for the BHOSLIB instances with an average gap of $-0.38\%$, and finds or improves the BKS for 53 out of 54 instances of the DIMACS class with an average gap of~$-0.13\%$.

\begin{figure}[htbp]
\hspace*{-0.6cm}
\includegraphics[width = 1.1\textwidth]{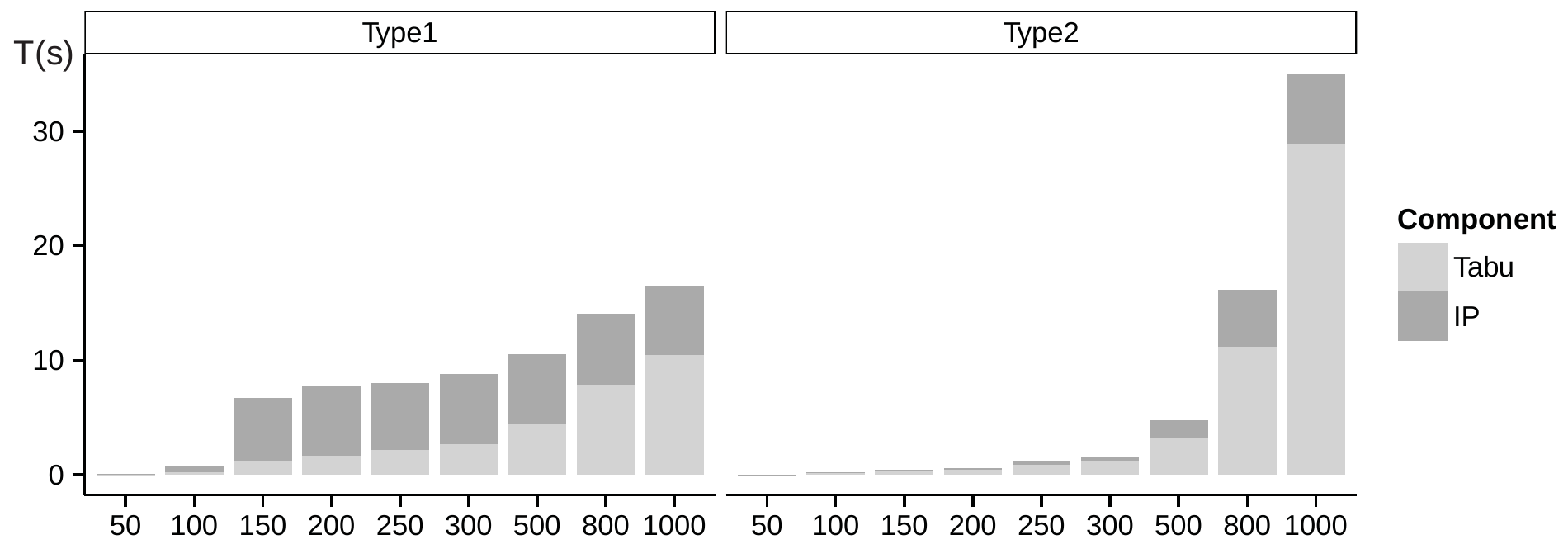}  \vspace*{-0.8cm}
\caption{Computational time of each component of HTS-DS as a function of $|V|$}
\label{fig:time}
\end{figure}

Moreover, since the scalability of the algorithm is essential for large scale applications, Figure~\ref{fig:time} presents a more detailed analysis of the CPU time spent in the two main phases of the method (tabu search and IP) as a function of the number of vertices $|V|$ for the instances of type T1 and T2. To eliminate some instances with few edges which tend to be easy to solve, we restricted this analysis to the subset of instances such that $|E| \geq 3 |V|$.

From this figure, we first observe that the average CPU time of HTS-DS remains below two seconds on a majority of instances. There are two situations where this computational time is exceeded. For the instances of type T1, the total time dedicated to the resolution of the reduced problems with the IP solver amounts to five to six seconds when $|V| \geq 150$ due to the increased difficulty of the mathematical models. For the instances of type T2, the resolution of the subproblems is faster, and the CPU time of the method only exceeds two seconds when $|V| \geq 500$.

Note that the time spent solving the subproblems cannot exceed an upper bound of 10 seconds overall due to the time limit imposed on the IP solver (1~second) and the limited number of subproblem resolutions (\mbox{$N_\textsc{Restart} = 10$}). Therefore, the scalability of the approach essentially depends on the efficiency of the tabu search, and more specifically, on the evaluation of the neighborhoods. For each instance type, we fitted the CPU time spent in the tabu search phase as a power law $f(|V|) = x |V|^y$ (by a least-squares regression of an affine function on the log-log graph). This time appears to grow as $\mathcal{O}(n^{1.81})$ on the instances of type T1, and  $\mathcal{O}(n^{2.30})$ on the instances of type T2.

 \subsection{Sensitivity analyses}
\label{sensitivity}

We performed sensitivity analyses in order to evaluate the impact of each main component and parameter of HTS-DS. Starting with the standard configuration described in Section~\ref{parameters}, we modified one parameter and design choice at a time (OFAT approach) to evaluate its effect. The following configurations were considered:
\begin{itemize}[nosep,leftmargin=0.5cm]
 \item[]\textbf{Standard.} Standard configuration described in Section~\ref{parameters}.
 \item[]\textbf{A. No Reduced Problem.} The reduced problem and IP solver is disabled.
 \item[]\textbf{B. No SWAP.} The SWAP neighborhood is not used.
 \item []\textbf{C. No Perturbation.} No perturbation: $\rho = 0$.
 \item []\textbf{D. $\uparrow$ Perturbation.} Higher level of perturbation: $\rho =0.4$.
 \item []\textbf{E. $\downarrow$ Tabu.} Shorter tabu tenure: $N_\textsc{tabu} =5$.
 \item []\textbf{F. $\uparrow$ Tabu.} Longer tabu tenure: $N_\textsc{tabu}=20$.
 \item []\textbf{G. $\downarrow$ Beta.} Shorter phases for penalty management: $\beta = 1.0$.
 \item []\textbf{H. $\uparrow$ Beta.}  Longer phases for penalty management: $\beta = 1.5$.
\end{itemize}
All configurations were run ten times on each instance. Table \ref{results:procedure} presents the gap of the best and average solutions over these runs, as well as the average CPU time resulting from each method configuration.

\begin{table}[htbp]
\centering
\renewcommand{\arraystretch}{1.2}
\setlength{\tabcolsep}{8pt}
\caption{Analysis of HTS-DS components}
\label{results:procedure}
 \scalebox{0.85}
 {
\begin{tabular}{|l|rrr|rrr|}
\cline{2-7}
\multicolumn{1}{l|}{} & \multicolumn{3}{c|}{\textbf{Type T1}} & \multicolumn{3}{c|}{\textbf{Type T2}} \\ \cline{2-7} 
 \multicolumn{1}{c|}{\strut}  &   \multicolumn{1}{c}{\textbf{$\text{Gap}_{\textsc{B}}$}} &\textbf{$\text{Gap}_{\textsc{A}}$} & \textbf{T(s)} & \textbf{$\text{Gap}_{\textsc{B}}$} & \textbf{$\text{Gap}_{\textsc{A}}$} & \textbf{T(s)} \\ \hline
Standard & \textbf{-0.04} & \textbf{0.00} & 6.18 & \textbf{0.00} & \textbf{0.01} & 4.41 \\   
A. No Reduced Problem & 0.06 & 0.13 & 2.74 & 0.09 & 0.12 & 3.56 \\
B. No SWAP & 0.07 & 0.14 & 5.10 & 0.01 & 0.02 & 3.36 \\ 
C. No Perturbation & 0.00 & 0.05 & 6.10 & 0.12 & 0.16 & 2.23 \\ 
D. $\uparrow$ Perturbation & -0.02 & 0.01 & 6.56 & 0.01 & 0.01 & 8.05 \\ 
E. $\downarrow$ Tabu & -0.01 & 0.02 & 6.13 & 0.01 & 0.02 & 4.94 \\ 
F. $\uparrow$ Tabu & 0.08 & 0.14 & 6.51 & 0.01 & 0.02 & 4.22 \\ 
G. $\downarrow$ Beta & 0.01 & 0.02 & 6.93 & 0.02 & 0.02 & 4.19 \\ 
H. $\uparrow$ Beta & 0.01 & 0.03 & 6.91 & 0.02 & 0.03 & 4.20 \\ \hline
\end{tabular}
}
\end{table}

These experiments (configurations A--C) highlight the major contribution of the mathematical programming solver used for the solution of the reduced problems, the limited SWAP neighborhood as well as the perturbation operator. Without the subproblem solver, the average gap from the BKS rises up to 0.13\% for type T1 and 0.12\% for type T2, while the CPU time decreases by 55\% for type T1 and 19\% for type T2. In a similar fashion, deactivating the SWAP or the perturbation operator translates into a significant decrease of solution quality for only a moderate reduction of CPU time.

The three main search parameters in charge of the perturbation rate, the tabu tenure and the management of the penalty factors also play an important role in the method. Increasing the perturbation rate to $\rho=0.4$, for example, allows to better diversify the search but leads to a loss of information from the best solution, with a negative impact on the overall performance (configuration D). Similarly, increasing the size of the tabu list is over-restrictive and hinders the progress towards high-quality solutions (configuration E), whereas decreasing it may increase the cycling probability (configuration F). Finally, the value of the parameter $\beta$ has been chosen so as to find trade-off solutions at the frontier of feasibility without spending too much time per phase. Increasing or decreasing this parameter (configurations G and H) leads to solutions of lower quality.

\section{Conclusions}
\label{sec:conclusion}

In this article, we have proposed a matheuristic \citep{Maniezzo2009} combining a tabu search with integer programing for the MWDS problem. The method exploits an efficient neighborhood search, an adaptive penalty scheme to explore intermediate infeasible solutions, perturbation mechanisms as well as additional intensification phases in which a reduced problem is optimized by means of an integer programming solver. The size of the reduced problem is adapted to fully exploit the capabilities of the solver.

Through extensive computational experiments, we have demonstrated the good performance of HTS-DS, which outperforms previous algorithms on the classical benchmark instances of \cite{Jovanovic2010,Wang2017a} with up to 4000 vertices. The method is efficient and scalable, with an observed CPU time growth in $\mathcal{O}(|V|^{1.81})$ and $\mathcal{O}(|V|^{2.30})$ as a function of the number of vertices for the instances of types T1 and T2, respectively, therefore making it suitable for large-scale applications. Finally, our sensitivity analyses demonstrate the essential contribution of each component of the search: the subproblem resolution, the perturbation mechanisms, and the restricted  SWAP neighborhood.

The research perspectives are numerous. First of all, we exploited the past search history and the \emph{frequency} of some vertices in the dominating set to fix variables in the subproblem. This strategy is closely related to the Construct, Merge, Solve {\&} Adapt (CMSA) approach described in \cite{Blum2016}. Moreover, other instance and solution metrics (e.g., ratio between weight and degree) may be used to further guide the search. Second, the synergies between heuristic search and mathematical-programming techniques can certainly be better exploited, by possibly sharing dual information or forming other types of subproblems. Finally, the current machine learning literature, and especially the study of graphs arising from social networks opens the way to very-large scale problems, with possibly millions of vertices, which deserve a careful study. Solution methods for such problems need to be carefully crafted to retain only essential search components working in linear or log-linear complexity. Overall, these are all open important research directions which can be explored in the near future.

\section*{Acknowledgments}

\noindent
This work  was partially supported by the \emph{Conselho Nacional de Desenvolvimento Cient\'ifico e Tecnol\'ogico} of Brasil (grant numbers 141225/2012, 308498/2015-1)  as well as \emph{Funda\c{c}\~ao Carlos Chagas Filho de Amparo \`a Pesquisa do Estado do Rio de Janeiro} (grant number E-26/203.310/2016). This support is gratefully acknowledged.

\section*{Appendix -- Detailed computational results}

Wang et al. \cite{Wang2017a} extended the BHOSLIB and DIMACS benchmark instances with vertices weights and made available some computational results. Tables \ref{tab:BHOSLIB} and \ref{tab:DIMACS} present detailed results on these instances, in the same format as Tables~\mbox{\ref{tab:MWDSt11} to \ref{tab:MWDSt22}}. Since some of the DIMACS instances are larger, we reduced the termination criterion of the method to $(I_{\textsc{max}},I_{\textsc{ni}})=(2000,1000)$ for this set.

\begin{table}[htpb] 
\renewcommand{\arraystretch}{1.2}
\setlength{\tabcolsep}{9pt}
\vspace*{-1cm}
\caption{Weighted BHOSLIB instances --  Comparison of HTS-DS with recent algorithms.}
 \label{tab:BHOSLIB}
\hspace*{-1cm}
 \scalebox{0.8}
 {
\begin{tabular}{|lrr|rr|rr|rrrrr|}
\hline
\multicolumn{3}{|c|}{\textbf{Instances}} & \multicolumn{2}{c|}{\textbf{ACO-PP-LS}$^{*}$} & \multicolumn{2}{c|}{\textbf{C${\text{C}}^2$FS}$^{*}$} & \multicolumn{5}{c|}{\textbf{HTS-DS}} \\ \hline
\textbf{Name} & \multicolumn{1}{c}{\textbf{$|V|$}} & \multicolumn{1}{c|}{\textbf{$|E|$}} & \multicolumn{1}{c}{\textbf{Best}} & \multicolumn{1}{c|}{\textbf{Avg}} & \multicolumn{1}{c}{\textbf{Best}} & \multicolumn{1}{c|}{\textbf{Avg}} & \multicolumn{1}{c}{\textbf{Best}} & \multicolumn{1}{c}{\textbf{Avg}} & \multicolumn{1}{c}{\textbf{T(s)}} & \multicolumn{1}{c}{\textbf{$\text{Gap}_{\textsc{B}}$}} & \multicolumn{1}{c|}{\textbf{$\text{Gap}_{\textsc{A}}$}} \\ \hline
frb30-15-1 & 450 & 17827 & 223 & 223.5 & 214 & 214 & \textbf{212} & \textbf{212.0} & 12.96 & \textbf{-0.93\%} & \textbf{-0.93\%} \\ 
frb30-15-2 & 450 & 17874 & 244 & 244 & \textbf{242} & \textbf{242.0} & \textbf{242} & \textbf{242.0} & 4.46 & 0\% & 0\% \\ 
frb30-15-3 & 450 & 17809 & 175 & 175 & \textbf{175} & \textbf{175.0} & \textbf{175} & \textbf{175.0} & 6.75 & 0\% & 0\% \\ 
frb30-15-4 & 450 & 17831 & 174 & 182.7 & \textbf{166} & 167 & \textbf{166} & \textbf{166.0} & 4.54 & 0\% & 0\% \\ 
frb30-15-5 & 450 & 17794 & 172 & 177.4 & \textbf{160} & \textbf{160.0} & \textbf{160} & \textbf{160.0} & 5.39 & 0\% & 0\% \\ 
frb35-17-1 & 595 & 27856 & 283 & 285.8 & \textbf{274} & \textbf{274.0} & \textbf{274} & \textbf{274.0} & 13.88 & 0\% & 0\% \\ 
frb35-17-2 & 595 & 27847 & 218 & 220.4 & \textbf{208} & \textbf{208.0} & \textbf{208} & \textbf{208.0} & 5.66 & 0\% & 0\% \\ 
frb35-17-3 & 595 & 27931 & 204 & 207 & \textbf{201} & \textbf{201.0} & \textbf{201} & \textbf{201.0} & 6.11 & 0\% & 0\% \\ 
frb35-17-4 & 595 & 27842 & 320 & 328.5 & 287 & 287 & \textbf{286} & \textbf{286.0} & 16.47 & \textbf{-0.35\%} & \textbf{-0.35\%} \\ 
frb35-17-5 & 595 & 28143 & 297 & 302.5 & \textbf{295} & 296.5 & \textbf{295} & \textbf{295.0} & 3.97 & 0\% & 0\% \\ 
frb40-19-1 & 760 & 41314 & 268 & 274.6 & \textbf{262} & \textbf{262.0} & \textbf{262} & \textbf{262.0} & 6.3 & 0\% & 0\% \\ 
frb40-19-2 & 760 & 41263 & 250 & 250.6 & \textbf{243} & 243.5 & \textbf{243} & \textbf{243.0} & 25.05 & 0\% & 0\% \\ 
frb40-19-3 & 760 & 41095 & 271 & 276.7 & 252 & 252 & \textbf{250} & \textbf{250.0} & 24.52 & \textbf{-0.79\%} & \textbf{-0.79\%} \\ 
frb40-19-4 & 760 & 41605 & 264 & 266.3 & 250 & 250 & \textbf{249} & 249.1 & 22.39 & \textbf{-0.40\%} & -0.36\% \\ 
frb40-19-5 & 760 & 41619 & 286 & 288.8 & \textbf{272} & 282.5 & \textbf{272} & \textbf{272.0} & 19.84 & 0\% & 0\% \\ 
frb45-21-1 & 945 & 59186 & 370 & 376.2 & \textbf{328} & 333.7 & \textbf{328} & \textbf{328.0} & 23.98 & 0\% & 0\% \\ 
frb45-21-2 & 945 & 58624 & 273 & 278.1 & \textbf{259} & 259.3 & \textbf{259} & 259.2 & 27.22 & 0\% & 0.08\% \\ 
frb45-21-3 & 945 & 58245 & 249 & 254.6 & \textbf{233} & 233.9 & \textbf{233} & \textbf{233.0} & 41.14 & 0\% & 0\% \\ 
frb45-21-4 & 945 & 58549 & 453 & 475.2 & \textbf{399} & \textbf{399.0} & \textbf{399} & \textbf{399.0} & 24.82 & 0\% & 0\% \\ 
frb45-21-5 & 945 & 58579 & 352 & 369.6 & 318 & 318.2 & \textbf{312} & 312.5 & 28.68 & \textbf{-1.89\%} & -1.73\% \\ 
frb50-23-1 & 1150 & 80072 & 293 & 298.9 & \textbf{261} & 267.8 & \textbf{261} & \textbf{261.0} & 49.65 & 0\% & 0\% \\ 
frb50-23-2 & 1150 & 80851 & 300 & 302.9 & \textbf{277} & \textbf{277.0} & \textbf{277} & \textbf{277.0} & 50.39 & 0\% & 0\% \\ 
frb50-23-3 & 1150 & 81068 & 313 & 315.6 & 297 & 298.1 & \textbf{281} & \textbf{281.0} & 39.22 & \textbf{-5.39\%} & \textbf{-5.39\%} \\ 
frb50-23-4 & 1150 & 80258 & 279 & 279 & \textbf{265} & \textbf{265.0} & \textbf{265} & \textbf{265.0} & 59.52 & 0\% & 0\% \\ 
frb50-23-5 & 1150 & 80035 & 445 & 445.4 & 415 & 421.4 & \textbf{404} & 408.3 & 33.8 & \textbf{-2.65\%} & -1.61\% \\ 
frb53-24-1 & 1272 & 94227 & 241 & 244 & \textbf{229} & \textbf{229.0} & \textbf{229} & \textbf{229.0} & 27.07 & 0\% & 0\% \\ 
frb53-24-2 & 1272 & 94289 & 318 & 318.8 & \textbf{298} & 300.3 & \textbf{298} & \textbf{298.0} & 108.45 & 0\% & 0\% \\ 
frb53-24-3 & 1272 & 94127 & 187 & 188.7 & \textbf{182} & \textbf{182.0} & \textbf{182} & \textbf{182.0} & 66.35 & 0\% & 0\% \\ 
frb53-24-4 & 1272 & 94308 & 202 & 202.4 & \textbf{189} & \textbf{189.0} & \textbf{189} & \textbf{189.0} & 88.61 & 0\% & 0\% \\ 
frb53-24-5 & 1272 & 94226 & 211 & 225.8 & \textbf{204} & \textbf{204.0} & \textbf{204} & \textbf{204.0} & 24.37 & 0\% & 0\% \\ 
frb56-25-1 & 1400 & 109676 & 231 & 231.9 & \textbf{229} & \textbf{229.0} & \textbf{229} & \textbf{229.0} & 33.54 & 0\% & 0\% \\ 
frb56-25-2 & 1400 & 109401 & 335 & 336 & \textbf{319} & \textbf{319.0} & \textbf{319} & \textbf{319.0} & 45.2 & 0\% & 0\% \\ 
frb56-25-3 & 1400 & 109379 & 346 & 351.5 & \textbf{336} & 343.1 & \textbf{336} & \textbf{336.0} & 50.85 & 0\% & 0\% \\ 
frb56-25-4 & 1400 & 110038 & 275 & 277.2 & 268 & 268 & \textbf{265} & \textbf{265.0} & 52.85 & \textbf{-1.12\%} & -1.12\% \\ 
frb56-25-5 & 1400 & 109601 & 495 & 498.9 & 426 & 429.7 & \textbf{408} & 411.4 & 39.96 & \textbf{-4.23\%} & -3.43\% \\ 
frb59-26-1 & 1534 & 126555 & 276 & 288.4 & \textbf{262} & 263.2 & \textbf{262} & 262.6 & 59.98 & 0\% & 0.23\% \\ 
frb59-26-2 & 1534 & 126163 & 426 & 426.1 & \textbf{383} & 388.8 & \textbf{383} & 386.6 & 49.61 & 0\% & 0.94\% \\ 
frb59-26-3 & 1534 & 126082 & 272 & 273.5 & 248 & 248 & \textbf{246} & 246.7 & 108.53 & \textbf{-0.81\%} & -0.52\% \\ 
frb59-26-4 & 1534 & 127011 & 256 & 265.3 & \textbf{248} & 248.1 & \textbf{248} & \textbf{248.0} & 111.37 & 0\% & 0\% \\ 
frb59-26-5 & 1534 & 125982 & 307 & 307.8 & 290 & 291.3 & \textbf{288} & 288.4 & 123.5 & \textbf{-0.69\%} & -0.55\% \\ 
frb100-40 & 4000 & 572774 & 377 & 384.2 & \textbf{350} & \textbf{350.0} & \textbf{350} & \textbf{350.0} & 300.18 & 0\% & 0\% \\ \hline
\multicolumn{3}{|r|}{\textbf{Avg}} & 286.12 & 290.73 & 268.63 & 270.01 & \textbf{267.07} & 267.41 & 45.05 & \textbf{-0.47\%} & \textbf{-0.38\%} \\ \hline
\multicolumn{10}{l}{* -- ACO-PP-LS and C${\text{C}}^2$FS were run until a fixed time limit of 1000s \cite{Wang2017a}} \vspace*{-0.15cm} \\
\multicolumn{10}{l}{\hspace*{0.53cm}(equivalent to 950s after CPU scaling)} \\
\end{tabular}
 }
\end{table}
 
\begin{table}[htpb] 
\renewcommand{\arraystretch}{1.065}
\setlength{\tabcolsep}{9pt}
\vspace*{-1.8cm}
\caption{Weighted DIMACS instances -- Comparison of HTS-DS with recent algorithms.}
 \label{tab:DIMACS}
\hspace*{-1cm}
 \scalebox{0.76}
 {
\begin{tabular}{|lrr|rr|rr|rrrrr|}
\hline
\multicolumn{3}{|c|}{\textbf{Instances}} & \multicolumn{2}{c|}{\textbf{ACO-PP-LS}$^{*}$} & \multicolumn{2}{c|}{\textbf{C${\text{C}}^2$FS}$^{*}$} & \multicolumn{5}{c|}{\textbf{HTS-DS}} \\ \hline
\textbf{Name} & \multicolumn{1}{c}{\textbf{$|V|$}} & \multicolumn{1}{c|}{\textbf{$|E|$}} & \multicolumn{1}{c}{\textbf{Best}} & \multicolumn{1}{c|}{\textbf{Avg}} & \multicolumn{1}{c}{\textbf{Best}} & \multicolumn{1}{c|}{\textbf{Avg}} & \multicolumn{1}{c}{\textbf{Best}} & \multicolumn{1}{c}{\textbf{Avg}} & \multicolumn{1}{c}{\textbf{T(s)}} & \multicolumn{1}{c}{\textbf{$\text{Gap}_{\textsc{B}}$}} & \multicolumn{1}{c|}{\textbf{$\text{Gap}_{\textsc{A}}$}} \\ \hline
brock200\_2 & 200 & 10024 & \textbf{23} & \textbf{23.0} & \textbf{23} & \textbf{23.0} & \textbf{23} & \textbf{23.0} & 0.14 & 0\% & 0\% \\ 
brock200\_4 & 200 & 6811 & \textbf{68} & 70.4 & \textbf{68} & \textbf{68.0} & \textbf{68} & \textbf{68.0} & 0.23 & 0\% & 0\% \\ 
brock400\_2 & 400 & 20014 & \textbf{65} & \textbf{65.0} & \textbf{65} & \textbf{65.0} & \textbf{65} & \textbf{65.0} & 0.91 & 0\% & 0\% \\ 
brock400\_4 & 400 & 20035 & \textbf{75} & 75.7 & \textbf{75} & \textbf{75.0} & \textbf{75} & \textbf{75.0} & 1.1 & 0\% & 0\% \\ 
brock800\_2 & 800 & 111434 & \textbf{28} & 28.4 & \textbf{28} & \textbf{28.0} & \textbf{28} & \textbf{28.0} & 5.19 & 0\% & 0\% \\ 
brock800\_4 & 800 & 111957 & \textbf{31} & 32.8 & \textbf{31} & \textbf{31.0} & \textbf{31} & \textbf{31.0} & 13.98 & 0\% & 0\% \\ 
C1000.9 & 1000 & 49421 & 197 & 197 & \textbf{191} & 194.8 & \textbf{191} & \textbf{191.0} & 14.87 & 0\% & 0\% \\ 
C2000.5 & 2000 & 999164 & \textbf{10} & \textbf{10.0} & \textbf{10} & \textbf{10.0} & \textbf{10} & \textbf{10.0} & 49.81 & 0\% & 0\% \\ 
C2000.9 & 2000 & 199468 & 136 & 139.3 & \textbf{130} & \textbf{130.0} & \textbf{130} & \textbf{130.0} & 38.85 & 0\% & 0\% \\ 
C250.9 & 250 & 3141 & \textbf{235} & \textbf{235.0} & \textbf{235} & \textbf{235.0} & \textbf{235} & \textbf{235.0} & 0.18 & 0\% & 0\% \\ 
C4000.5 & 500 & 12418 & \textbf{9} & \textbf{9.0} & \textbf{9} & \textbf{9.0} & \textbf{9} & \textbf{9.0} & 214.9 & 0\% & 0\% \\ 
C500.9 & 200 & 1534 & \textbf{226} & \textbf{226.0} & 228 & 228 & \textbf{226} & \textbf{226.0} & 4.03 & 0\% & 0\% \\ 
c-fat200-1 & 200 & 1534 & 232 & 232.9 & \textbf{226} & \textbf{226.0} & \textbf{226} & \textbf{226.0} & 0.05 & 0\% & 0\% \\ 
c-fat200-2 & 200 & 3235 & \textbf{57} & \textbf{57.0} & \textbf{57} & \textbf{57.0} & \textbf{57} & \textbf{57.0} & 0.07 & 0\% & 0\% \\ 
c-fat200-5 & 200 & 8473 & \textbf{9} & \textbf{9.0} & \textbf{9} & \textbf{9.0} & \textbf{9} & \textbf{9.0} & 0.08 & 0\% & 0\% \\ 
c-fat500-1 & 500 & 4459 & 568 & 568 & \textbf{522} & \textbf{522.0} & \textbf{522} & \textbf{522.0} & 0.31 & 0\% & 0\% \\ 
c-fat500-2 & 500 & 9139 & 283 & 283 & \textbf{261} & \textbf{261.0} & \textbf{261} & \textbf{261.0} & 0.34 & 0\% & 0\% \\ 
c-fat500-5 & 500 & 23191 & \textbf{20} & 20.8 & \textbf{20} & \textbf{20.0} & \textbf{20} & \textbf{20.0} & 0.52 & 0\% & 0\% \\ 
DSJC1000\_5 & 1000 & 499652 & 14 & 14.2 & 14 & 14 & \textbf{13} & \textbf{13.0} & 17.56 & \textbf{-7.14\%} & \textbf{-7.14\%} \\ 
DSJC500\_5 & 500 & 125248 & \textbf{15} & \textbf{15.0} & \textbf{15} & \textbf{15.0} & 16 & 16 & 1.95 & 6.67\% & 6.67\% \\ 
gen200\_p0.9\_44 & 200 & 1990 & \textbf{458} & \textbf{458.0} & 470 & 470 & \textbf{458} & \textbf{458.0} & 2.54 & 0\% & 0\% \\ 
gen200\_p0.9\_55 & 200 & 1990 & \textbf{433} & 439.7 & \textbf{433} & \textbf{433.0} & \textbf{433} & \textbf{433.0} & 0.15 & 0\% & 0\% \\ 
gen400\_p0.9\_55 & 400 & 7980 & 293 & 303.6 & 288 & 288 & \textbf{284} & \textbf{284.0} & 0.2 & \textbf{-1.39\%} & \textbf{-1.39\%} \\ 
gen400\_p0.9\_65 & 400 & 7980 & 291 & 291.2 & \textbf{287} & \textbf{287.0} & \textbf{287} & \textbf{287.0} & 0.86 & 0\% & 0\% \\ 
gen400\_p0.9\_75 & 400 & 7980 & \textbf{307} & \textbf{307.0} & \textbf{307} & \textbf{307.0} & \textbf{307} & \textbf{307.0} & 1.15 & 0\% & 0\% \\ 
hamming10-4 & 1024 & 89600 & 88 & 88 & \textbf{86} & \textbf{86.0} & \textbf{86} & \textbf{86.0} & 0.37 & 0\% & 0\% \\ 
hamming8-2 & 256 & 1024 & 1744 & 1748.2 & 1744 & 1744 & \textbf{1737} & \textbf{1737.0} & 5.24 & \textbf{-0.40\%} & \textbf{-0.40\%} \\ 
hamming8-4 & 256 & 11776 & 74 & 76.5 & 71 & 71 & \textbf{68} & \textbf{68.0} & 0.28 & \textbf{-4.23\%} & \textbf{-4.23\%} \\ 
johnson32-2-4 & 496 & 14880 & \textbf{192} & 192.1 & \textbf{192} & \textbf{192.0} & \textbf{192} & \textbf{192.0} & 0.53 & 0\% & 0\% \\ 
keller4 & 171 & 5100 & 228 & 233.1 & \textbf{220} & \textbf{220.0} & \textbf{220} & \textbf{220.0} & 1.07 & 0\% & 0\% \\ 
keller5 & 776 & 74710 & 189 & 196.7 & 182 & 182 & \textbf{181} & \textbf{181.0} & 0.34 & \textbf{-0.55\%} & \textbf{-0.55\%} \\ 
keller6 & 3361 & 1026582 & 81 & 82.4 & \textbf{80} & \textbf{80.0} & \textbf{80} & \textbf{80.0} & 12.12 & 0\% & 0\% \\ 
MANN\_a27 & 378 & 702 & \textbf{405} & \textbf{405.0} & \textbf{405} & \textbf{405.0} & \textbf{405} & \textbf{405.0} & 0.13 & 0\% & 0\% \\ 
MANN\_a45 & 1035 & 1980 & \textbf{1080} & \textbf{1080.0} & \textbf{1080} & \textbf{1080.0} & \textbf{1080} & \textbf{1080.0} & 0.84 & 0\% & 0\% \\ 
MANN\_a81 & 3321 & 6480 & \textbf{3402} & \textbf{3402.0} & \textbf{3402} & \textbf{3402.0} & \textbf{3402} & \textbf{3402.0} & 8.58 & 0\% & 0\% \\ 
p\_hat1500-1.clq & 1500 & 284923 & 68 & 68 & \textbf{56} & \textbf{56.0} & \textbf{56} & \textbf{56.0} & 32.04 & 0\% & 0\% \\ 
p\_hat1500-2.clq & 1500 & 568960 & \textbf{14} & \textbf{14.0} & \textbf{14} & \textbf{14.0} & \textbf{14} & \textbf{14.0} & 30.51 & 0\% & 0\% \\ 
p\_hat1500-3.clq & 1500 & 847244 & 6 & 6 & \textbf{5} & \textbf{5.0} & \textbf{5} & \textbf{5.0} & 31.24 & 0\% & 0\% \\ 
p\_hat300-1.clq & 300 & 10933 & 104 & 104.9 & \textbf{99} & 99.6 & \textbf{99} & \textbf{99.0} & 0.53 & 0\% & 0\% \\ 
p\_hat300-2.clq & 300 & 21928 & \textbf{31} & \textbf{31.0} & \textbf{31} & \textbf{31.0} & \textbf{31} & \textbf{31.0} & 0.43 & 0\% & 0\% \\ 
p\_hat300-3.clq & 300 & 33390 & \textbf{8} & \textbf{8.0} & \textbf{8} & \textbf{8.0} & \textbf{8} & \textbf{8.0} & 0.48 & 0\% & 0\% \\ 
p\_hat700-1.clq & 700 & 60999 & 76 & 76 & \textbf{67} & \textbf{67.0} & \textbf{67} & \textbf{67.0} & 3.43 & 0\% & 0\% \\ 
p\_hat700-2.clq & 700 & 121728 & \textbf{21} & \textbf{21.0} & \textbf{21} & \textbf{21.0} & \textbf{21} & \textbf{21.0} & 2.98 & 0\% & 0\% \\ 
p\_hat700-3.clq & 700 & 183010 & \textbf{6} & \textbf{6.0} & \textbf{6} & \textbf{6.0} & \textbf{6} & \textbf{6.0} & 3.21 & 0\% & 0\% \\ 
san1000 & 1000 & 250500 & \textbf{8} & \textbf{8.0} & \textbf{8} & \textbf{8.0} & \textbf{8} & \textbf{8.0} & 55.31 & 0\% & 0\% \\ 
san200\_0.7\_1 & 200 & 5970 & 81 & 83.6 & \textbf{73} & \textbf{73.0} & \textbf{73} & \textbf{73.0} & 0.17 & 0\% & 0\% \\ 
san200\_0.7\_2 & 200 & 5970 & \textbf{53} & \textbf{53.0} & \textbf{53} & \textbf{53.0} & \textbf{53} & \textbf{53.0} & 0.17 & 0\% & 0\% \\ 
san200\_0.9\_1 & 200 & 1990 & \textbf{368} & \textbf{368.0} & \textbf{368} & \textbf{368.0} & \textbf{368} & \textbf{368.0} & 0.04 & 0\% & 0\% \\ 
san200\_0.9\_2 & 200 & 1990 & \textbf{406} & 406.4 & \textbf{406} & \textbf{406.0} & \textbf{406} & \textbf{406.0} & 0.06 & 0\% & 0\% \\ 
san200\_0.9\_3 & 200 & 1990 & \textbf{328} & \textbf{328.0} & \textbf{328} & \textbf{328.0} & \textbf{328} & \textbf{328.0} & 0.05 & 0\% & 0\% \\ 
san400\_0.5\_1 & 400 & 39900 & \textbf{16} & \textbf{16.0} & \textbf{16} & \textbf{16.0} & \textbf{16} & \textbf{16.0} & 0.73 & 0\% & 0\% \\ 
san400\_0.7\_1 & 400 & 23940 & \textbf{44} & \textbf{44.0} & \textbf{44} & \textbf{44.0} & \textbf{44} & \textbf{44.0} & 0.71 & 0\% & 0\% \\ 
san400\_0.7\_2 & 400 & 23940 & \textbf{42} & \textbf{42.0} & \textbf{42} & \textbf{42.0} & \textbf{42} & \textbf{42.0} & 0.56 & 0\% & 0\% \\ 
san400\_0.7\_3 & 400 & 23940 & \textbf{40} & \textbf{40.0} & \textbf{40} & \textbf{40.0} & \textbf{40} & \textbf{40.0} & 0.76 & 0\% & 0\% \\ \hline
\multicolumn{3}{|r|}{\textbf{Avg}} & 246.04 & 247.02 & 243.5 & 243.58 & \textbf{242.96} & \textbf{242.96} & 10.42 & \textbf{-0.13\%} & \textbf{-0.13\%} \\ \hline    
\multicolumn{10}{l}{* -- ACO-PP-LS and C${\text{C}}^2$FS were run until a fixed time limit of 1000s \cite{Wang2017a}} \vspace*{-0.15cm} \\
\multicolumn{10}{l}{\hspace*{0.53cm}(equivalent to 950s after CPU scaling)} \\
\end{tabular}
 }
\end{table}

\end{document}